\documentclass[]{unireason}
\usepackage{fix-cm}
\usepackage{helvet}           

\usepackage{nicefrac}         
\usepackage{siunitx}         

\usepackage{longtable}
\usepackage{booktabs}

\usepackage{tabularx}         
\usepackage{array}           
\usepackage{makecell}         
\usepackage{threeparttable}

\usepackage{wrapfig}          
\usepackage{float}            
\usepackage[export]{adjustbox}

\usepackage{tikz}             
\usepackage{pgfplots}         
\usepackage{pgf-pie}

\usepackage{listings}        
\usepackage{ragged2e}        
\usepackage{comment}

\usepackage{pifont}           
\usepackage{fontawesome}

\usepackage{enumitem}        
\usepackage{titletoc}         
\usepackage{minitoc}          
\usepackage[toc,page,header]{appendix}  
\usepackage{fontawesome}
\usepackage{url}              
\usepackage[hang,flushmargin]{footmisc}  
\pgfplotsset{compat=1.18}

\usepackage{xspace}

\newcommand{\modelname}{\textbf{DeepGen 1.0}\xspace}

\definecolor{lightblue}{RGB}{200, 230, 255}  
\definecolor{headerblue}{RGB}{150, 200, 255}

\usepackage{xcolor}

\usepackage{multirow}
\usepackage{xcolor,colortbl}
\definecolor{lavendergray}{rgb}{0.77, 0.76, 0.82}
\definecolor{lightgray}{rgb}{0.83, 0.83, 0.83}

\usepackage{array}
\newcolumntype{H}{>{\setbox0=\hbox\bgroup}c<{\egroup}@{}}
\newcommand{\tablestyle}[2]{\setlength{\tabcolsep}{#1}\renewcommand{\arraystretch}{#2}\centering\small}

\newcommand{\myparagraph}[1]{\vspace{0pt}\noindent{\bf #1}}

\usepackage{amssymb}%
\usepackage{pifont}%

\usepackage{parskip}

\usepackage{wrapfig}

\definecolor{myblue}{rgb}{0.11764705882352941, 0.5647058823529412, 1.0}
\definecolor{Gray}{gray}{0.9}
\definecolor{darkgreen}{rgb}{0.545, 0.749, 0.608}

\usepackage{tablefootnote}

\usepackage{soul}
\sethlcolor{Gray}

\newcounter{examplebox}



\title{%
  \begin{minipage}[c]{0.08\textwidth}
    \includegraphics[height=2.8em]{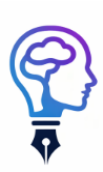} 
  \end{minipage}%
  \hspace{0.1em}%
  \begin{minipage}[c]{0.90\textwidth}
    \centering
    \textbf{\modelname}: A Lightweight Unified Multimodal Model for Advancing Image Generation and Editing
  \end{minipage}
}

\author{
    Dianyi Wang\textsuperscript{1,2*$\dagger$}, 
    Ruihang Li\textsuperscript{1,3*}
    Feng Han\textsuperscript{1,2*}, 
    Chaofan Ma\textsuperscript{4*}, 
    Wei Song\textsuperscript{1,5,6*}, \\
    Siyuan Wang\textsuperscript{8*}, 
    Yibin Wang\textsuperscript{1,2*}, 
    Yi Xin\textsuperscript{1,7}, 
    Hongjian Liu\textsuperscript{3}, 
    Zhixiong Zhang\textsuperscript{1,4}, \\
    Shengyuan Ding\textsuperscript{1,2}, 
    Tianhang Wang\textsuperscript{1,5}, 
    Zhenglin Cheng\textsuperscript{1,5,6},
    Tao Lin\textsuperscript{6},
    Cheng Jin\textsuperscript{2}, \\
    Kaicheng Yu\textsuperscript{6},
    Jingjing Chen\textsuperscript{2},
    Wenjie Wang\textsuperscript{3},
    Zhongyu Wei\textsuperscript{1,2},
    Jiaqi Wang\textsuperscript{1$\dagger$}
}

\affiliation[1]{\mbox{Shanghai Innovation Institute}}
\affiliation[2]{\mbox{Fudan University}}
\affiliation[3]{\mbox{University of Science and Technology of China}}
\affiliation[4]{\mbox{Shanghai Jiao Tong University}}
\affiliation[5]{\mbox{Zhejiang University}}
\affiliation[6]{\mbox{Westlake University}}
\affiliation[7]{\mbox{Nanjing University}}
\affiliation[8]{\mbox{University of Southern California}}

\contribution[\textbf{*}]{Equal Contribution, \textsuperscript{$\dagger$}Project Leaders
}

\titlefigure{
  \begin{center}
    \includegraphics[width=0.95\linewidth]{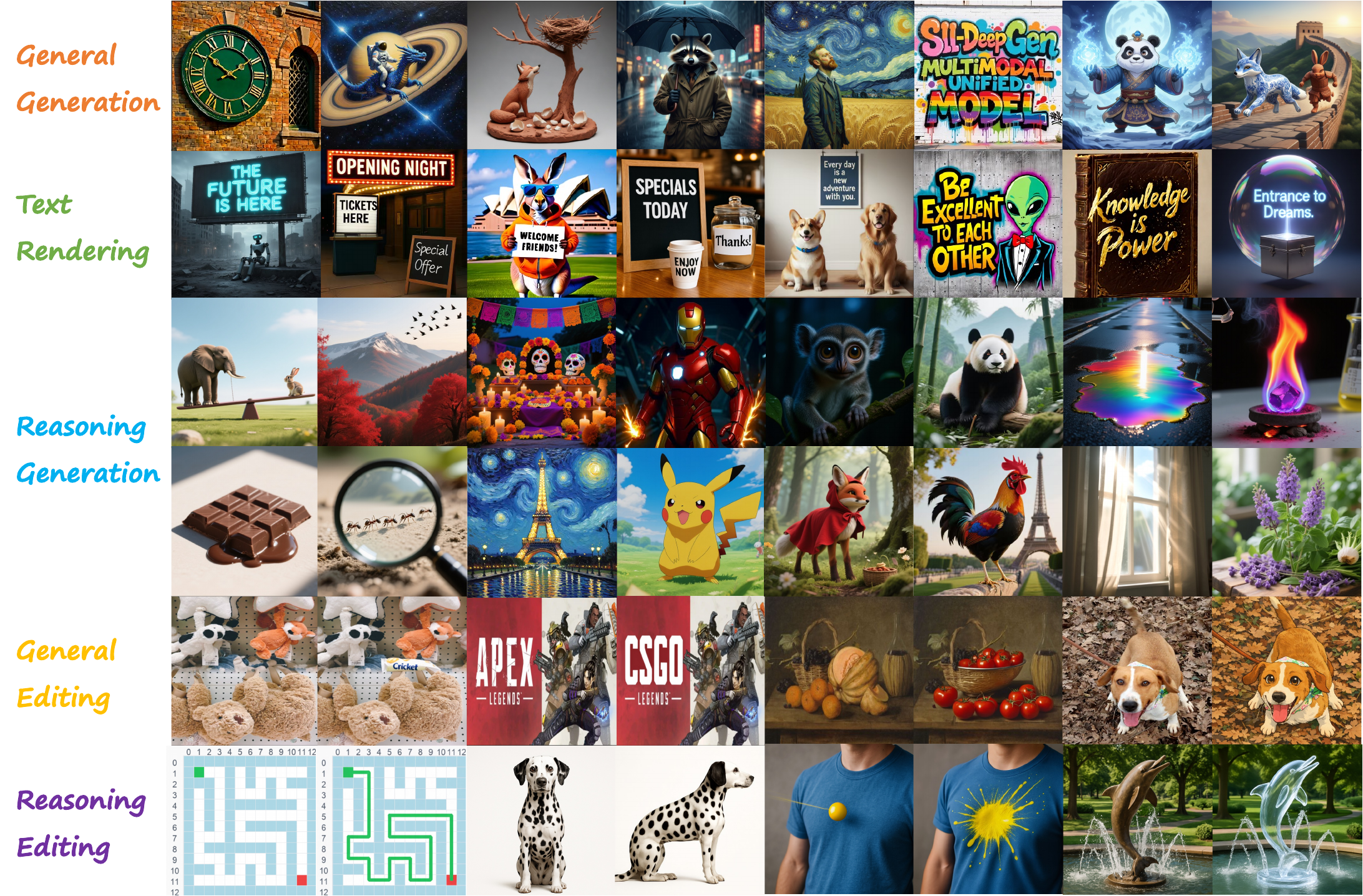}
    \captionof{figure}{Overview of DeepGen 1.0’s visual generation and editing abilities, including reasoning-intensive scenarios.}
    \label{Fig:teaser}
    \label{fig:teaser}
  \end{center}
}

\abstract{
    Current unified multimodal models for image generation and editing typically rely on massive parameter scales (e.g., $>$10B), entailing prohibitive training costs and deployment footprints. 
    In this work, we present \modelname, a lightweight 5B unified model that achieves comprehensive capabilities competitive with or surpassing much larger counterparts. 
    To overcome the limitations of compact models in semantic understanding and fine-grained control, we introduce \textbf{Stacked Channel Bridging (SCB)}, a deep alignment framework that extracts hierarchical features from multiple VLM layers and fuses them with learnable ``think tokens'' to provide the generative backbone with structured, reasoning-rich guidance. 
    We further design a data-centric training strategy spanning three progressive stages: (1) \textbf{Alignment Pre-training} on large-scale image-text pairs and editing triplets to synchronize VLM and DiT representations, (2) \textbf{Joint Supervised Fine-tuning} on a high-quality mixture of generation, editing, and reasoning tasks to foster omni-capabilities, and (3) \textbf{Reinforcement Learning with MR-GRPO}, which leverages a mixture of reward functions and supervision signals, resulting in substantial gains in generation quality and alignment with human preferences, while maintaining stable training progress and avoiding visual artifacts.
    Despite being trained on only $\sim$50M samples, \modelname achieves leading performance across diverse benchmarks, surpassing the 80B HunyuanImage by 28\% on WISE and the 27B Qwen-Image-Edit by 37\% on UniREditBench.
    By open-sourcing our training code, weights, and datasets, we provide an efficient, high-performance alternative to democratize unified multimodal research.
}

\checkdata[GitHub]{\url{https://github.com/DeepGenTeam/DeepGen}}
\checkdata[HuggingFace]{\url{https://huggingface.co/DeepGenTeam/DeepGen-1.0}}
\checkdata[Datasets]{\url{https://huggingface.co/datasets/DeepGenTeam/DeepGen-1.0}}

\begin{document}
\maketitle


\vspace{1.5em}

\section{Introduction}

\begin{figure*}[t]
    \centering
    \includegraphics[width=0.9\linewidth]{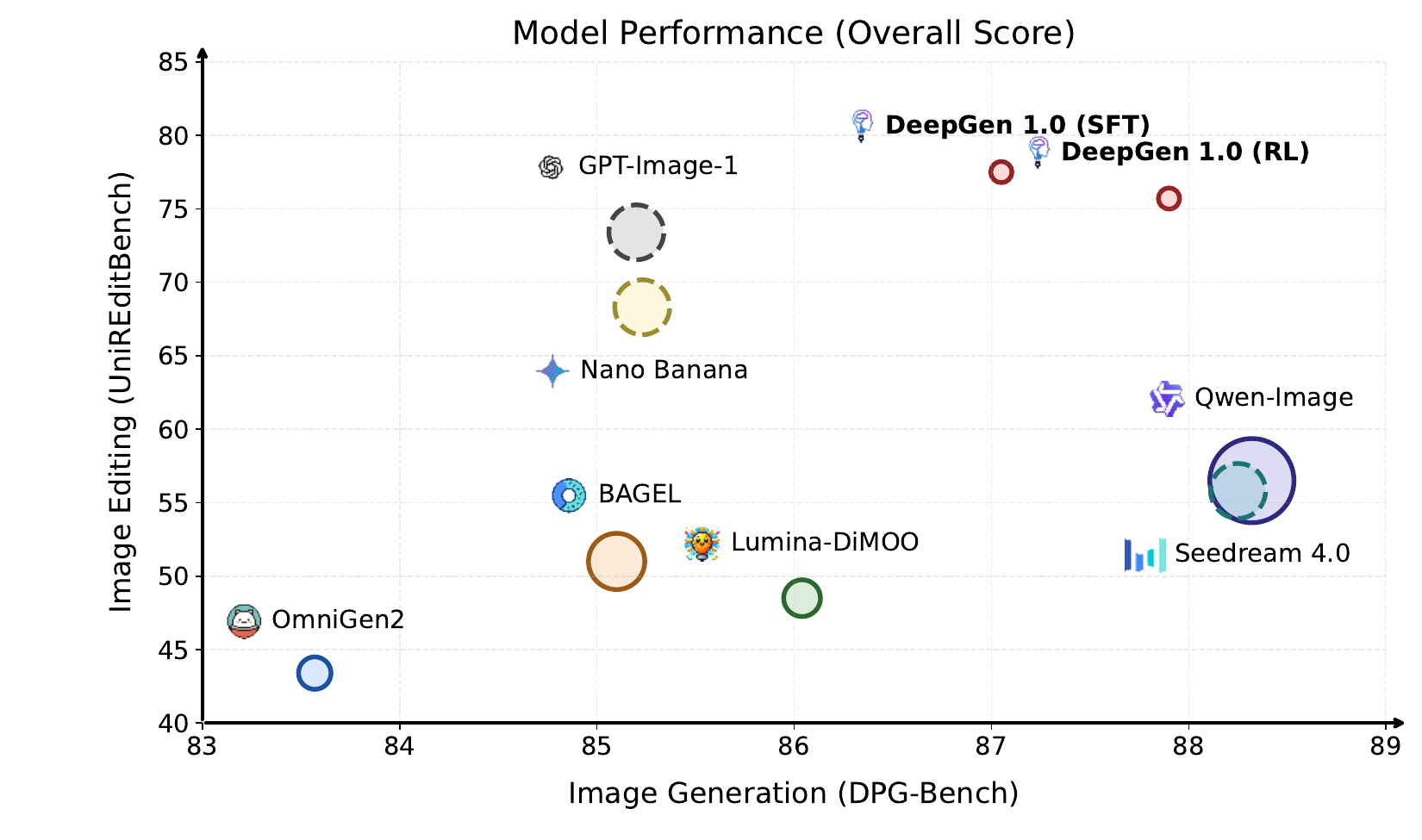} 
    \caption{Model performance comparison on image generation and editing benchmarks. Bubble size is proportional to model parameter count. Dashed outer rings indicate models with unreported parameter counts. Higher scores correspond to better performance.}
    \label{Fig:performance_comparison}
\end{figure*}

Advancing image generation and editing to handle increasingly complex instructions requires models that go beyond mere pixel synthesis to possess deep semantic understanding. 
To meet this demand, a promising paradigm has emerged that integrates the comprehensive capabilities of vision-language models (VLMs) with the generative power of diffusion models, aiming to achieve semantically accurate generation and precise editing. 
Closed-source systems such as GPT-Image-1~\cite{ OpenAIGPTImage1} and Nano Banana~\cite{google2025gemini25flashimage} have validated this potential. 
In the open-source domain, a recent wave of models, including BAGEL~\cite{deng2025emergingpropertiesunifiedmultimodal}, HunyuanImage~3.0~\cite{cao2025hunyuanimage30technicalreport}, Qwen-Image~\cite{wu2025qwenimagetechnicalreport}, and LongCat-Image~\cite{meituanlongcatteam2025longcatimagetechnicalreport}, has actively explored this direction to elevate generative performance through unified understanding. 
These advancements underscore the transformative impact of unified models in redefining the boundaries of visual generation.

Despite this rapid progress, current high-performing unified models remain prohibitively \textit{expensive}.
Models such as Qwen-Image (27B), HunyuanImage~3.0 (80B), BAGEL (14B), and Emu3.5 (34B) all demand billions of training samples and massive computational resources.
Many further require separate generation and editing models, doubling the total parameter count, \textit{e.g.}, pushing deployment footprints to a total of 54B for Qwen-Image \& Qwen-Image-Edit and 26B for LongCat-Image \& LongCat-Image-Edit.
While the need for lightweight alternatives is clear, existing small-scale unified models~\cite{chen2025blip3ofamilyfullyopen,xie2025showosingletransformerunify,ma2025janusflow} have consistently underperformed across diverse tasks, thereby reinforcing a common perception: compact models lack the capacity for comprehensive multimodal generation and editing.
\textit{Interestingly}, a closer examination of recent benchmarks challenges this view: performance does not scale monotonically with model size.
For example, as shown in Fig.~\ref{Fig:performance_comparison}, Lumina-DiMOO (8B) achieves a generation score of 86.04 on DPG-Bench, surpassing the larger BAGEL (14B, 85.10). 
Similar patterns are observed across other benchmarks and evaluation dimensions (Table~\ref{table:general}, \ref{table:wise}, \ref{table:corebench}, \ref{table:reason_edit}, and \ref{table:text}). 
This indicates that, for unified multimodal models, larger scale \textit{alone} does not necessarily guarantee stronger performance.

Motivated by this observation, we argue that \textit{\emph{a lightweight model, when empowered by synergistic architecture design and data-centric training strategies, can achieve comprehensive capabilities competitive with or even surpassing much larger counterparts}}.
To substantiate this, we present \modelname, a compact framework with a total of \textit{5B} parameters (3B VLM and 2B DiT) that integrates general generation, reasoning generation, text rendering, general editing, and reasoning editing within a \textit{single} model.
Despite its compact size, \modelname achieves results competitive with or exceeding models 3$\times$ to 16$\times$ its size, as highlighted in Fig.~\ref{Fig:performance_comparison}.
For instance, in general instruction following DPG-Bench, \modelname attains 87.90, eclipsing massive baselines like HunyuanImage~3.0 (86.10). Moving to reasoning-intensive tasks, it achieves 0.73 on WISE, outperforming the 80B HunyuanImage~3.0 (0.57) by a remarkable \textbf{28\%} margin. Furthermore, on the editing front, it dominates the UniREditBench with 77.5, surpassing the dedicated 27B Qwen-Image-Edit (56.5) by over \textbf{37\%}. Across the board, \modelname demonstrates that intelligent design can triumph over raw scale.
Remarkably, the entire training requires only $\sim$50M samples across a simple three-stage pipeline, compared to 1.2B samples for LongCat-Image and 5B for HunyuanImage~3.0.

To support these comprehensive capabilities within a compact 5B budget, we introduce a specialized architecture that maximizes VLM-DiT synergy. 
\modelname employs a 3B VLM~\cite{bai2025qwen25vltechnicalreport} as the understanding and reasoning backbone and a 2B DiT~\cite{wei2026skyworkunipic20building} as the generative backbone.
To align these two modules, we propose \textbf{Stacked Channel Bridging (SCB)}. 
SCB first extracts hidden states from \textit{six uniformly distributed} VLM layers (spanning low, mid, and high levels) to capture hierarchical features from visual and text inputs. 
To further enhance reasoning, we inject learnable ``think tokens'' that act as an implicit chain of thoughts. 
These multi-source features are then \textit{channel-wise concatenated} and fused via a lightweight \textit{connector} into a dense multimodal conditional sequence. 
Unlike prior methods that rely on the final VLM layer~\cite{wu2025qwenimagetechnicalreport,wu2025omnigen2explorationadvancedmultimodal} or use average pooling~\cite{shen2025mammothmoda2} that blurs fine-grained details, this design fully preserves both fine-grained visual details and high-level semantics, while providing the DiT with structured, reasoning-rich guidance.

\begin{figure*}[t]
    \centering
    \includegraphics[width=\linewidth]{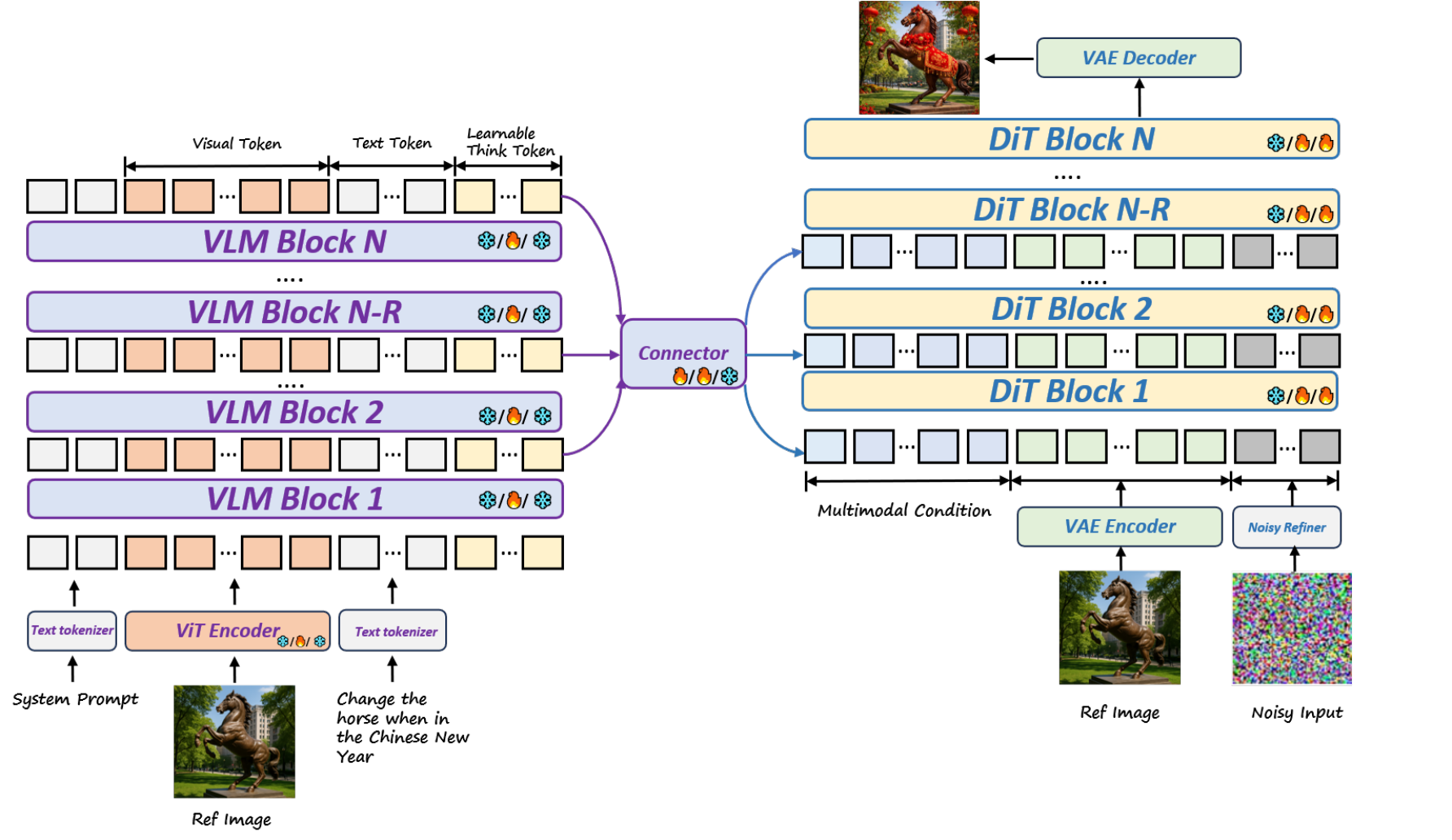} 
    \vspace{-10pt}
    \caption{Overview of \textbf{DeepGen 1.0} architecture. DeepGen 1.0 adopts a unified VLM-DiT paradigm with a dual-branch visual encoding strategy: a ViT encoder captures high-level semantics for the VLM, while a VAE encoder extracts compressed latent features for the DiT. Multimodal conditions derived from the VLM, together with reference-image VAE latents, are concatenated with the target image’s noise tokens to form a single DiT input sequence, enabling self-attention over both conditioning and generation signals. Stacked channel bridging (SCB) performs deep feature fusion between the VLM and DiT to strengthen generation and editing, while DiT positional encodings explicitly distinguish reference tokens from target tokens. Icons shown at the right of each block indicate whether the corresponding module is frozen or trainable during the Pre-Training, SFT, and RL stages, respectively.}
    \vspace{-2mm}
    \label{Fig:arch}
\end{figure*}

To fully unlock the potential of of \modelname's compact architecture, we design a data-centric training strategy tailored for tight VLM-DiT integration in the low-parameter regime. 
This strategy emphasizes simplicity and data efficiency across three progressively stages. 
First, in \textbf{Alignment Pre-training}, we optimize only the connector and learnable think tokens to align VLM representations with the DiT's latent space, utilizing large-scale image-text pairs and editing triplets.
Second, during \textbf{Joint Supervised Fine-tuning (SFT)}, we unfreeze the DiT and apply LoRA to the VLM for end-to-end optimization. 
We curate a high-quality data mixture by integrating general generation and editing data, reasoning-based generation and editing data, and text-rendering data to foster omni-capabilities while preserving the VLM's inherent knowledge.
Finally, we employ \textbf{Reinforcement Learning (RL)} to further align the model with human preferences. 
We adopt our novel MR-GRPO, with mixture of rewards and supervision signals, enhancing it with decoupled advantage normalization~\cite{liu2026gdpo} to better preserve multi-reward granularity. 
To prevent capability degradation during RL, we introduce an auxiliary supervised diffusion loss, ensuring the model retains the broad capabilities acquired during the {joint supervised fine-tuning} stage.

\begin{figure*}[t]
    \centering
    \includegraphics[width=0.92\linewidth]{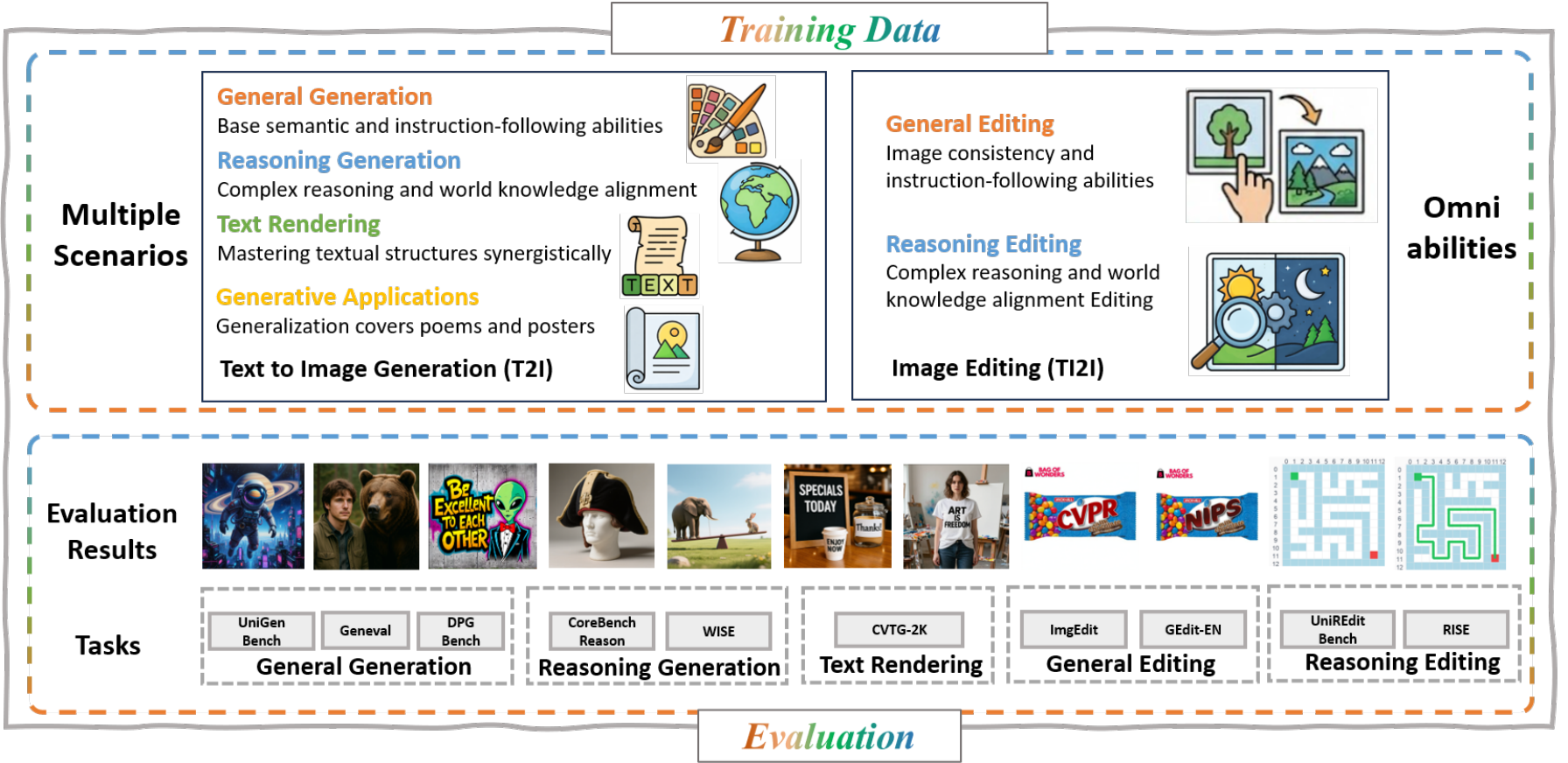} 
    \caption{Overview of our training data for broad omni-capabilities and comprehensive evaluation across benchmarks.}
    \label{Fig:data}
\end{figure*}

Our contributions are summarized as follows:
\begin{itemize}
\item We present \modelname, a compact 5B unified model that integrates general generation, reasoning, text rendering, and editing within a single framework. Despite its small size, it achieves performance competitive with or surpassing models up to 16$\times$ larger (\textit{e.g.}, 80B), demonstrating that massive scaling is not the sole path to high-performance multimodal generation.

\item We propose Stacked Channel Bridging (SCB), a lightweight alignment module that fuses multi-layer VLM features via channel concatenation and a shallow connector. Augmented with learnable think tokens, SCB enables deep semantic transfer from the VLM to the DiT while preserving fine-grained visual details, offering a superior alternative to standard final-layer or average-pooling approaches.

\item We design a data-centric training strategy spanning three progressive stages: (1) alignment pre-training on large-scale pairs and triplets, (2) joint SFT on a high-quality mixture of generation, reasoning, editing, and text rendering tasks, and (3) we propose MR-GRPO for RL alignment with auxiliary supervision and mixture of rewards , enabling stable preference optimization without capability degradation.

\item We conduct comprehensive evaluations across diverse benchmarks, demonstrating leading performance among open-source models in reasoning-based generation and editing, while maintaining competitive general generation quality.

\item We publicly release the \modelname framework, including model weights, training and evaluation code, and key data components. By providing an efficient and high-performance alternative to resource-intensive large models, we aim to democratize unified multimodal research and empower broader community exploration.
\end{itemize}

\section{Model Architecture}

DeepGen 1.0 follows a VLM-DiT architecture as shown in Fig~\ref{Fig:arch}, where the VLM offers strong multimodal understanding with well cross-modal alignment and rich world knowledge to capture complex multimodal priors from both textual and visual inputs. The DiT serves as a high-fidelity generation decoder guided by multimodal conditional inputs extracted from the VLM. We utilize Qwen-2.5-VL (3B)~\cite{bai2025qwen25vltechnicalreport} as our pretrained VLM and SD3.5-Medium (2B) as our DiT, initialized from~\cite{wei2026skyworkunipic20building} with joint generation and editing capability. Feature alignment is achieved via a streamlined connector module, which instantiates a SigLIP visual encoder~\cite{zhai2023sigmoidlosslanguageimage} followed by six transformer layers~\cite{wu2025openuni}. This compact design maintains a total model size of approximately 5B parameters, striking an optimal balance between performance and computational efficiency.

\paragraph{\textbf{Stacked Channel Bridging (SCB)}} Prior unified multimodal models~\cite{wu2025qwenimagetechnicalreport,wu2025omnigen2explorationadvancedmultimodal,meituanlongcatteam2025longcatimagetechnicalreport,lin2025uniworldv1highresolutionsemanticencoders} typically take the final-layer (or penultimate-layer) hidden states of a VLM, transform them through a connector, and use them as multimodal conditional input to the DiT. This design has two key limitations. First, the final VLM layers are heavily biased toward high-level semantic abstraction, often discarding fine-grained visual details that are critical for DiT modeling~\cite{li2025unifusionvisionlanguagemodelunified}. Second, relying on a single layer makes the conditional signal vulnerable to layer-specific representation biases, which can hinder stable alignment and effective fusion between the VLM and DiT. 
An alternative line of work~\cite{deng2025emergingpropertiesunifiedmultimodal,wang2025lightfusionlightweighteddoublefusion,shi2024lmfusion} performs deep fusion by introducing shared attention between the VLM and DiT at every layer. However, this approach substantially increases parameter scale and optimization complexity, making efficient and reliable training challenging. 
Subsequent works~\cite{shen2025mammothmoda2} aggregate hidden states from multiple VLM layers using average pooling. 

To more effectively and efficiently aggregate features from multiple VLM layers while preserving fine-grained information and enhancing reasoning, we propose the Stacked Channel Bridging (SCB) framework. SCB operates through three integrated steps:

- \textbf{Think Token Injection.} While standard VLM representations provide rich interleaved multimodal signals~\cite{pan2025transfer,chen2025blip3ofamilyfullyopen}, explicit reasoning tokens can further act as implicit Chains of Thought (CoT). To strengthen the model’s reasoning capability, we first inject a fixed set of learnable ``think tokens'' into the VLM input sequence. These tokens interact with textual and visual inputs across all layers via self-attention, progressively summarizing hidden representations and effectively extracting knowledge encoded in the VLM.

\begin{table*}[t]
\centering
\caption{Comparison of different models across general image generation and editing benchmarks. Top-1/2/3 results within each column excluding closed-source models are marked with gold, silver, and bronze icons.
}
\tablestyle{1pt}{1.1}
\setlength\tabcolsep{7pt}
\resizebox{0.9\textwidth}{!}{
\begin{tabular}{ccccccc}
\toprule
\multicolumn{1}{c|}{}                         & \multicolumn{1}{c|}{}                         & \multicolumn{3}{c|}{\bf General T2I Generation}                                                             & \multicolumn{2}{c}{\bf General Editing} \\
\multicolumn{1}{c|}{\multirow{-2}{*}{\bf Model}}  & \multicolumn{1}{c|}{\multirow{-2}{*}{\bf Params}} & \textbf{GenEval↑}       & \textbf{DPGBench↑}                                             & \multicolumn{1}{c|}{\bf UniGenBench↑} & \textbf{ImgEdit↑}          & \textbf{GEdit-EN↑}         \\ \midrule
\rowcolor[HTML]{F6D6D3}\multicolumn{7}{c}{Closed-source Models}                                                                                                                                                                                                      \\ \midrule
\multicolumn{1}{c|}{Nano Banana}               & \multicolumn{1}{c|}{–}                        & 0.75          & 85.23                                                & \multicolumn{1}{c|}{87.45}       & 4.35             & 7.54             \\
\multicolumn{1}{c|}{GPT-Image-1}              & \multicolumn{1}{c|}{–}                        & 0.84          & 85.20                                                & \multicolumn{1}{c|}{92.77}       & 4.20             & 7.53             \\
\multicolumn{1}{c|}{Seedream 4.0}              & \multicolumn{1}{c|}{–}                        & 0.84          & \cellcolor[HTML]{FFFFFF}{\color[HTML]{1F2328} 88.25} & \multicolumn{1}{c|}{87.30}       & 4.18             & 7.68             \\
\multicolumn{1}{c|}{FLUX.1 Kontext {[}Pro{]}} & \multicolumn{1}{c|}{–}                        & –             & \cellcolor[HTML]{FFFFFF}{\color[HTML]{1F2328} –}     & \multicolumn{1}{c|}{75.84}       & 4.00             & 6.56             \\ \midrule
\rowcolor[HTML]{DCEBFA}\multicolumn{7}{c}{Open-source Models}                                                                                                                                                                                                        \\ \midrule
\multicolumn{1}{c|}{Janus-Pro}                & \multicolumn{1}{c|}{7B}                       & 0.80          & 84.20                                                & \multicolumn{1}{c|}{61.61}       & –                & –                \\
\multicolumn{1}{c|}{Show-o2}                  & \multicolumn{1}{c|}{7B}                       & 0.76          & \cellcolor[HTML]{FFFFFF}{\color[HTML]{1F2328} 86.14} & \multicolumn{1}{c|}{62.73}       & –                & –                \\
\multicolumn{1}{c|}{BLIP3-o}                  & \multicolumn{1}{c|}{7B + 1.4B}                & 0.84          & \cellcolor[HTML]{FFFFFF}{\color[HTML]{1F2328} 81.60} & \multicolumn{1}{c|}{59.87}       & –                & –                \\
\multicolumn{1}{c|}{MetaQuery-XL}             & \multicolumn{1}{c|}{7B+ 1.6B}                 & 0.80          & \cellcolor[HTML]{FFFFFF}{\color[HTML]{1F2328} 82.05} & \multicolumn{1}{c|}{–}           & –                & –                \\
\multicolumn{1}{c|}{OmniGen2}                & \multicolumn{1}{c|}{3B + 4B}                  & 0.80          & \cellcolor[HTML]{FFFFFF}{\color[HTML]{1F2328} 83.57} & \multicolumn{1}{c|}{63.09}       & 3.43             & 6.41             \\
\multicolumn{1}{c|}{UniWorld v1}              & \multicolumn{1}{c|}{7B + 12B}                 & 0.80          & \cellcolor[HTML]{FFFFFF}{\color[HTML]{1F2328} 81.38} & \multicolumn{1}{c|}{63.11}       & 3.26             & 4.85             \\
\multicolumn{1}{c|}{BAGEL}                    & \multicolumn{1}{c|}{14B}                      & 0.82          & 85.10                                                & \multicolumn{1}{c|}{61.53}       & 3.20             & 6.52             \\
\multicolumn{1}{c|}{FLUX.1 {[}Dev{]}}         & \multicolumn{1}{c|}{12B}                      & 0.82          & 83.84                                                & \multicolumn{1}{c|}{69.88}       & –                & –                \\
\multicolumn{1}{c|}{X-Omni}                   & \multicolumn{1}{c|}{7B + 12B}                 & 0.83          & 87.65\includegraphics[width=1em]{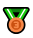}                                                & \multicolumn{1}{c|}{53.77}       & –                & –                \\
\multicolumn{1}{c|}{Lumina-DiMOO}             & \multicolumn{1}{c|}{8B}                       & 0.88\includegraphics[width=1em]{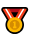}          & 86.04                                                & \multicolumn{1}{c|}{71.12}       & –                & –                \\
\multicolumn{1}{c|}{Mammoth2}                 & \multicolumn{1}{c|}{8B + 3B + 2B}             & 0.87\includegraphics[width=1em]{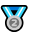}          & 87.20                                                & \multicolumn{1}{c|}{–}           & 4.06             & 6.60             \\
\multicolumn{1}{c|}{LongCat-Image}            & \multicolumn{1}{c|}{7B + 6B}                    & 0.87\includegraphics[width=1em]{images/silver.png}          & 86.80                                                & \multicolumn{1}{c|}{–}           & –                & –                \\
\multicolumn{1}{c|}{LongCat-Image-Edit}       & \multicolumn{1}{c|}{7B + 6B}                    & –             & –                                                    & \multicolumn{1}{c|}{–}           & 4.50\includegraphics[width=1em]{images/gold.png}             & 7.60\includegraphics[width=1em]{images/gold.png}              \\
\multicolumn{1}{c|}{Hunyuan-Image 3.0}        & \multicolumn{1}{c|}{80B}                      & 0.72          & 86.10                                                & \multicolumn{1}{c|}{–}           & –                & –                \\
\multicolumn{1}{c|}{Z-Image-Turbo}            & \multicolumn{1}{c|}{4B + 6B}                  & 0.84          & 85.15                                                & \multicolumn{1}{c|}{71.40}       & –                & –                \\
\multicolumn{1}{c|}{Qwen-Image}               & \multicolumn{1}{c|}{7B + 20B}                 & 0.87\includegraphics[width=1em]{images/silver.png}          & 88.32\includegraphics[width=1em]{images/gold.png}                                                & \multicolumn{1}{c|}{78.81\includegraphics[width=1em]{images/gold.png}}       & –                & –                \\
\multicolumn{1}{c|}{Qwen-Image-Edit [2509]}          & \multicolumn{1}{c|}{7B + 20B}                 & –             & –                                                    & \multicolumn{1}{c|}{–}           & 4.35\includegraphics[width=1em]{images/silver.png}             & 7.54\includegraphics[width=1em]{images/silver.png}              \\
\multicolumn{1}{c|}{GLM-Image}                & \multicolumn{1}{c|}{9B + 7B}                    & –             & 84.78                                                & \multicolumn{1}{c|}{–}           & –                & –                \\ \midrule
\multicolumn{1}{c|}{\bf DeepGen 1.0 (SFT)}              & \multicolumn{1}{c|}{\textbf{3B + 2B}}           & 0.86\includegraphics[width=1em]{images/bronze.png}  & 87.05                                                & \multicolumn{1}{c|}{74.18\includegraphics[width=1em]{images/bronze.png} }       & 4.09             & 7.12             \\
\multicolumn{1}{c|}{\bf DeepGen 1.0 (RL)}           & \multicolumn{1}{c|}{\textbf{3B + 2B}}           & 0.87\includegraphics[width=1em]{images/silver.png}          & 87.90\includegraphics[width=1em]{images/silver.png}                                                & \multicolumn{1}{c|}{75.74\includegraphics[width=1em]{images/silver.png}}       & 4.14\includegraphics[width=1em]{images/bronze.png}             & 7.17\includegraphics[width=1em]{images/bronze.png}             \\ \midrule
\end{tabular}}
\label{table:general}
\end{table*}

- \textbf{Layer Selection.} With the think tokens injected, we select multiple VLM hidden states to fuse, balancing performance and computational efficiency. Instead of relying on a single layer, and following~\cite{wang2025activating} which suggests that sparsely and uniformly distributed layers within VLMs provide effective representations for visual information, we select six hidden states sampled uniformly across the low-, mid-, and high-level layers. This ensures the capture of varying-granularity visual features and semantics, alongside the reasoning information embedded in the think token positions.

- \textbf{Feature Fusion.} Finally, we integrate the selected multi-layer hidden states, which now encode both multimodal features and think token representations. Given a set of selected VLM hidden states $[x_{1}, \dots, x_{n}] \in \mathbb{R}^{L \times d}$ where $n$ denotes the number of selected layers and $L$ is the sequence length (including think tokens), we first stack them along the channel dimension. This concatenated feature tensor in dimension $d'$ is then projected to match the DiT input width using a lightweight two-layer MLP. The aligned features are then fed into a Transformer-encoder-based connector to deeply fuse information across layers, producing the final robust conditional input $c\in \mathbb{R}^{L \times d_{\text{DiT}}}$:
\begin{equation}
c = \text{Encoder}(\text{MLP}(\mathrm{Concat}_{\text{ch}}(x_{1}, \dots, x_{n}))).
\end{equation}

\begin{table*}[t]
\centering
\caption{Evaluation of reasoning-based text-to-image generation involving world knowledge on the WISE~\cite{niu2025wiseworldknowledgeinformedsemantic} benchmark. "*" denotes generation with textual CoT reasoning.}
\tablestyle{1pt}{1.1}
\setlength\tabcolsep{9pt}
\resizebox{1.0\textwidth}{!}{
\begin{tabular}{ccccccccc}
\toprule
\multicolumn{1}{c|}{\bf Model}                               & \multicolumn{1}{c|}{\bf Params}                   & \textbf{Cultural}                    & \textbf{Time}                        & \textbf{Space}                       & \textbf{Biology}                     & \textbf{Physics}                     & \multicolumn{1}{c|}{\bf Chemistry}                   & \textbf{Overall↑}                    \\ \midrule
\rowcolor[HTML]{F6D6D3}\multicolumn{9}{c}{Closed-source Models}                                                                                                                                                                                                                                                                                                        \\ \midrule
\multicolumn{1}{c|}{{\color[HTML]{1F2329} GPT-Image-1}}       & \multicolumn{1}{c|}{{\color[HTML]{1F2329} –}} & {\color[HTML]{1F2329} 0.81} & {\color[HTML]{1F2329} 0.71} & {\color[HTML]{1F2329} 0.89} & {\color[HTML]{1F2329} 0.83} & {\color[HTML]{1F2329} 0.79} & \multicolumn{1}{c|}{{\color[HTML]{1F2329} 0.74}} & {\color[HTML]{1F2329} 0.80} \\
\multicolumn{1}{c|}{{\color[HTML]{1F2329} Seedream 4.0}} & \multicolumn{1}{c|}{{\color[HTML]{1F2329} –}} & {\color[HTML]{1F2329} 0.78} & {\color[HTML]{1F2329} 0.73} & {\color[HTML]{1F2329} 0.85} & {\color[HTML]{1F2329} 0.79} & {\color[HTML]{1F2329} 0.84} & \multicolumn{1}{c|}{{\color[HTML]{1F2329} 0.67}} & {\color[HTML]{1F2329} 0.78} \\ \midrule
\rowcolor[HTML]{DCEBFA}\multicolumn{9}{c}{Open-source Models}                                                                                                                                                                                                                                                                                   \\ \midrule
\multicolumn{1}{c|}{Janus-Pro}                           & \multicolumn{1}{c|}{7B}                       & 0.30                        & 0.37                        & 0.49                        & 0.36                        & 0.42                        & \multicolumn{1}{c|}{0.26}                        & 0.35                        \\
\multicolumn{1}{c|}{FLUX.1 {[}Dev{]}}                    & \multicolumn{1}{c|}{12B}                      & 0.48                        & 0.58                        & 0.62                        & 0.42                        & 0.51                        & \multicolumn{1}{c|}{0.35}                        & 0.50                        \\
\multicolumn{1}{c|}{MetaQuery-XL}                        & \multicolumn{1}{c|}{7B+ 1.6B}                 & 0.56                        & 0.55                        & 0.62                        & 0.49                        & 0.63                        & \multicolumn{1}{c|}{0.41}                        & 0.55                        \\
\multicolumn{1}{c|}{BLIP3-o}                             & \multicolumn{1}{c|}{7B + 1.4B}                & –                           & –                           & –                           & –                           & –                           & \multicolumn{1}{c|}{–}                           & 0.62                        \\
\multicolumn{1}{c|}{UniWorld-V1}                         & \multicolumn{1}{c|}{7B + 12B}                 & 0.53                        & 0.55                        & 0.73                        & 0.45                        & 0.59                        & \multicolumn{1}{c|}{0.41}                        & 0.55                        \\
\multicolumn{1}{c|}{OmniGen2}                            & \multicolumn{1}{c|}{3B + 4B}                  & 0.42                        & 0.52                        & 0.64                        & 0.43                        & 0.50                        & \multicolumn{1}{c|}{0.34}                        & 0.47                        \\
\multicolumn{1}{c|}{BAGEL\textsuperscript{*}}                               & \multicolumn{1}{c|}{14B}                      & 0.76                        & 0.69                        & 0.75                        & 0.65                        & 0.75                        & \multicolumn{1}{c|}{0.58}                        & 0.70\includegraphics[width=1em]{images/bronze.png}                        \\
\multicolumn{1}{c|}{NextFlow-RL}                         & \multicolumn{1}{c|}{7B + 18B}                 & 0.63                        & 0.63                        & 0.77                        & 0.58                        & 0.67                        & \multicolumn{1}{c|}{0.39}                        & 0.62                        \\
\multicolumn{1}{c|}{STAR}                                & \multicolumn{1}{c|}{7B}                       & 0.61                        & 0.67                        & 0.61                        & 0.74                        & 0.69                        & \multicolumn{1}{c|}{0.66}                        & 0.66                        \\
\multicolumn{1}{c|}{Hunyuan-Image 3.0}                   & \multicolumn{1}{c|}{80B}                      & 0.58                        & 0.57                        & 0.70                        & 0.56                        & 0.63                        & \multicolumn{1}{c|}{0.31}                        & 0.57                        \\
\multicolumn{1}{c|}{Qwen-Image}                          & \multicolumn{1}{c|}{7B + 20B}                 & 0.62                        & 0.63                        & 0.77                        & 0.57                        & 0.75                        & \multicolumn{1}{c|}{0.40}                        & 0.62                        \\
\multicolumn{1}{c|}{LongCat-Image}                       & \multicolumn{1}{c|}{7B + 6B}                  & 0.66                        & 0.61                        & 0.72                        & 0.66                        & 0.72                        & \multicolumn{1}{c|}{0.49}                        & 0.65                        \\ \midrule
\multicolumn{1}{c|}{\bf DeepGen 1.0 (SFT)}                             & \multicolumn{1}{c|}{\textbf{3B + 2B}}                  & 0.70                        & 0.71                        & 0.82                        & 0.62                        & 0.79                        & \multicolumn{1}{c|}{0.65}                        & 0.72\includegraphics[width=1em]{images/silver.png}                        \\
\multicolumn{1}{c|}{\bf DeepGen 1.0 (RL)}                          & \multicolumn{1}{c|}{\textbf{3B + 2B}}                  & 0.72                        & 0.81                        & 0.70                        & 0.67                        & 0.82                        & \multicolumn{1}{c|}{0.66}                        & 0.73\includegraphics[width=1em]{images/gold.png}                        \\ \midrule
\end{tabular}}
\label{table:wise}
\end{table*}

\section{Training}

\subsection{Stage 1: Alignment Pre-Training}
In the initial stage, we focus on establishing alignment between the VLM and the DiT. To achieve this, we train only the connector and 128 learnable think tokens while keeping all other model parameters frozen.
This phase utilizes general text-to-image generation and image editing tasks. Specifically, the model is trained for 200,000 iterations with the data details listed in Table~\ref{tab:data_details}. All images are generated at a fixed resolution of $512 \times 512$. We utilize a learning rate of $1 \times 10^{-4}$ with 20,000 warm-up steps. For a complete list of hyperparameters, please refer to Table \ref{tab:hyperparameters} in Appendix \ref{pretraing_sft_details}.

\subsection{Stage 2: Joint Supervised Fine-Tuning}
In the second stage, we unfreeze the entire model and conduct a joint VLM-DiT training, aiming to strengthen instruction-following capability and image synthesis quality with improved visual fidelity, semantic alignment, and knowledge-aware reasoning. To mitigate potential degradation of the VLM’s multimodal comprehension during joint optimization, we apply LoRA~\cite{hu2022lora} for efficient fine-tuning of the VLM.
We train the model on a diverse and high-quality mixture of tasks designed to foster omni abilities, including general text-to-image generation and editing, reasoning-based generation and editing, and text rendering. 


We perform supervised fine-tuning for 400,000 iterations on the multi-task dataset detailed in table~\ref{tab:data_details}. Images are trained at a fixed resolution of $512 \times 512$ while preserving the original aspect ratio via dynamic resizing. The model is optimized with a learning rate of $5 \times 10^{-5}$ with 20,000 warm-up steps. Detailed LoRA configurations and hyperparameters are provided in Table~\ref{tab:hyperparameters} of Appendix~\ref{pretraing_sft_details}.

DeepGen 1.0 follows a VLM-DiT architecture as shown in Fig~\ref{Fig:arch}, where the VLM offers strong multimodal understanding with well cross-modal alignment and rich world knowledge to capture complex multimodal priors from both textual and visual inputs. The DiT serves as a high-fidelity generation decoder gudided by multimodal conditional inputs extracted from the VLM. We utilize Qwen-2.5-VL (3B)~\cite{bai2025qwen25vltechnicalreport} as our pretrained VLM and SD3.5-Medium (2B) as our DiT, initialized from~\cite{wei2026skyworkunipic20building} with joint generation–editing capability. Feature alignment is achieved via a streamlined connector module, which instantiates a SigLIP visual encoder~\cite{zhai2023sigmoidlosslanguageimage} followed by six transformer layers~\cite{wu2025openuni}. This compact design maintains a total model size of approximately 5B parameters, striking an optimal balance between performance and computational efficiency.

\begin{table*}[t]
\centering
\caption{Evaluation of reasoning-based text-to-image generation with the philosophical framework on the T2I-CoREBench~\cite{li2025easierpaintingthinkingtexttoimage} benchmark through Qwen3-VL-32B-Thinking~\cite{bai2025qwen30vl}. "*" denotes generation with textual CoT reasoning.}
\tablestyle{12pt}{1.1}
\setlength\tabcolsep{8pt}
\resizebox{1.0\textwidth}{!}{
\begin{tabular}{ccccccccccc}
\toprule
\multicolumn{1}{c|}{\bf Model}                               & \multicolumn{1}{c|}{\bf Params}                   & \bf R-LR                        & \bf R-BR                        & \bf R-HR                        & \bf R-PR                        & \bf R-GR                        & \bf R-AR                        & \bf R-CR                        & \multicolumn{1}{c|}{\bf R-RR} & \bf Overall↑ \\ \midrule
\rowcolor[HTML]{F6D6D3}\multicolumn{11}{c}{Closed-source Models}                                                                                                                                                                                                                                                                                                                         \\ \midrule
\multicolumn{1}{c|}{{\color[HTML]{1F2329} Nano Banana}}  & \multicolumn{1}{c|}{{\color[HTML]{1F2329} –}} & {\color[HTML]{1F2329} 65.4} & {\color[HTML]{1F2329} 59.7} & {\color[HTML]{1F2329} 57.2} & {\color[HTML]{1F2329} 88.3} & {\color[HTML]{1F2329} 83.5} & {\color[HTML]{1F2329} 84.1} & {\color[HTML]{1F2329} 67.5} & \multicolumn{1}{c|}{58.7} & 70.5     \\
\multicolumn{1}{c|}{{\color[HTML]{1F2329} GPT-Image-1}}  & \multicolumn{1}{c|}{{\color[HTML]{1F2329} –}} & {\color[HTML]{1F2329} 61.6} & {\color[HTML]{1F2329} 52.0} & {\color[HTML]{1F2329} 58.1} & {\color[HTML]{1F2329} 89.9} & {\color[HTML]{1F2329} 76.7} & {\color[HTML]{1F2329} 82.4} & {\color[HTML]{1F2329} 67.7} & \multicolumn{1}{c|}{47.5} & 67.0     \\
\multicolumn{1}{c|}{{\color[HTML]{1F2329} Seedream 4.0}} & \multicolumn{1}{c|}{{\color[HTML]{1F2329} –}} & {\color[HTML]{1F2329} 79.2} & {\color[HTML]{1F2329} 51.4} & {\color[HTML]{1F2329} 52.9} & {\color[HTML]{1F2329} 89.1} & {\color[HTML]{1F2329} 88.6} & {\color[HTML]{1F2329} 80.1} & {\color[HTML]{1F2329} 70.8} & \multicolumn{1}{c|}{42.8} & 69.4     \\ \midrule
\rowcolor[HTML]{DCEBFA}\multicolumn{11}{c}{Open-source Models}                                                                                                                                                                                                                                                                                                    \\ \midrule
\multicolumn{1}{c|}{Janus-Pro}                           & \multicolumn{1}{c|}{7B}                       & 27.2                        & 15.9                        & 28.0                        & 25.4                        & 7.3                         & 30.8                        & 8.8                         & \multicolumn{1}{c|}{4.6}  & 18.5     \\
\multicolumn{1}{c|}{FLUX.1 {[}Dev{]}}                    & \multicolumn{1}{c|}{12B}                      & 26.3                        & 18.0                        & 25.9                        & 66.8                        & 38.0                        & 59.7                        & 35.7                        & \multicolumn{1}{c|}{18.1} & 36.1     \\
\multicolumn{1}{c|}{Show-o2}                             & \multicolumn{1}{c|}{7B}                       & 30.2                        & 21.3                        & 29.4                        & 59.7                        & 40.4                        & 54.7                        & 32.8                        & \multicolumn{1}{c|}{13.1} & 35.2     \\
\multicolumn{1}{c|}{BLIP3-o}                             & \multicolumn{1}{c|}{7B + 1.4B}                & 18.4                        & 16.0                        & 19.0                        & 44.6                        & 45.0                        & 51.1                        & 36.8                        & \multicolumn{1}{c|}{12.3} & 30.4     \\
\multicolumn{1}{c|}{OmniGen2}                            & \multicolumn{1}{c|}{3B + 4B}                  & 26.8                        & 19.2                        & 32.9                        & 64.1                        & 37.5                        & 56.5                        & 37.9                        & \multicolumn{1}{c|}{13.6} & 36.1     \\
\multicolumn{1}{c|}{BAGEL\textsuperscript{*}}                               & \multicolumn{1}{c|}{14B}                      & 28.6                        & 22.2                        & 24.8                        & 66.2                        & 55.8                        & 59.5                        & 42.6                        & \multicolumn{1}{c|}{29.3} & 41.1     \\
\multicolumn{1}{c|}{Hunyuan-Image 3.0}                   & \multicolumn{1}{c|}{80B}                      & 41.6                        & 27.4                        & 42.3                        & 76.3                        & 52.7                        & 52.2                        & 55.1                        & \multicolumn{1}{c|}{20.6} & 46.0     \\
\multicolumn{1}{c|}{Qwen-Image}                          & \multicolumn{1}{c|}{7B + 20B}                 & 42.2                        & 29.5                        & 40.0                        & 78.6                        & 47.9                        & 55.2                        & 59.0                        & \multicolumn{1}{c|}{18.4} & 46.3\includegraphics[width=1em]{images/bronze.png}     \\
\multicolumn{1}{c|}{Z-Image-Turbo}                          & \multicolumn{1}{c|}{4B + 6B}                 & 37.8                        & 24.8                        & 37.8                        & 75.6                        & 46.0                        & 59.4                        & 49.6                        & \multicolumn{1}{c|}{18.6} & 43.7     \\
\multicolumn{1}{c|}{LongCat-Image}                       & \multicolumn{1}{c|}{7B + 6B}                  & 41.7                        & 32.2                        & 38.4                        & 78.3                        & 72.6                        & 66.3                        & 55.8                        & \multicolumn{1}{c|}{32.6} & 52.2\includegraphics[width=1em]{images/gold.png}     \\ \midrule
\multicolumn{1}{c|}{\bf DeepGen 1.0 (SFT)}                             & \multicolumn{1}{c|}{\textbf{3B + 2B}}                  & 38.8                        & 28.7                        & 40.2                        & 79.1                        & 51.5                        & 65.7                        & 42.0                        & \multicolumn{1}{c|}{19.8} & 45.7     \\
\multicolumn{1}{c|}{\bf DeepGen 1.0 (RL)}                          & \multicolumn{1}{c|}{\textbf{3B + 2B}}                  & 38.5                        & 29.0                        & 41.2                        & 79.5                        & 51.9                        & 66.9                        & 45.6                        & \multicolumn{1}{c|}{19.6} & 46.5\includegraphics[width=1em]{images/silver.png}     \\ \midrule
\end{tabular}}
\label{table:corebench}
\end{table*}

\subsection{Stage 3: Reinforcement Learning}
To further improve generation quality and alignment with human preferences, we apply reinforcement learning after supervised fine-tuning. We propose the MR-GRPO framework, a variant of Pref-GRPO~\cite{wang2025pref}, which extends Group Relative Policy Optimization (GRPO)~\cite{shao2024deepseekmath} to flow matching models by performing on-policy stochastic sampling and evaluating each generated image with a mixture of pointwise and pairwise reward models. We further introduce a novel auxiliary supervised diffusion loss that complements KL regularization to mitigate capability degradation during prolonged RL training. In addition, we validate and adopt two concurrent improvements into our pipeline: (1) a noise-preserving stochastic sampling strategy~\cite{wang2025coefficients} that produces cleaner samples and more accurate reward signals, and (2) a decoupled advantage normalization scheme~\cite{liu2026gdpo} that better preserves multi-reward signal granularity.

Concretely, given a text condition $h$, the flow model samples a group of $G$ images $\{x^i_0\}_{i=1}^G$ and the corresponding denoising trajectories $\{x^i_T, x^i_{T-1}, \ldots, x^i_0\}_{i=1}^G$. For multi-reward optimization with reward functions $\{R_k\}_{k=1}^{K}$, we normalize each reward independently within each group before aggregation, following~\cite{liu2026gdpo}:
\begin{equation}
A^{i}_k = \frac{R_k(x^i_0, h) - \operatorname{mean}(\{R_k(x^j_0, h)\}_{j=1}^G)}{\operatorname{std}(\{R_k(x^j_0, h)\}_{j=1}^G)},
\end{equation}
and obtain the final advantage $\hat{A}^{i}$ via weighted aggregation $\sum_k w_k A^{i}_k$ followed by batch-wise normalization across the training batch. The training objective is:
\begin{equation}
\mathcal{L}_{\text{GRPO}}(\theta) = \mathbb{E}_{h \sim \mathcal{D}} \left[\frac{1}{G}\sum_{i=1}^{G}\frac{1}{T}\sum_{t=0}^{T-1} \left(\min\!\left(r^i_t(\theta)\, \hat{A}^i,\; \operatorname{clip}(r^i_t(\theta), 1\!-\!\epsilon, 1\!+\!\epsilon)\, \hat{A}^i\right) - \beta\, D_{\text{KL}}(\pi_\theta \| \pi_{\text{ref}})\right)\right],
\end{equation}
where $r^i_t(\theta) = {p_\theta(x^i_{t-\Delta t}|x^i_t, h)}\big/{p_{\theta_{\text{old}}}(x^i_{t-\Delta t}|x^i_t, h)}$ is the per-step importance ratio. We use 3 complementary reward functions to jointly optimize visual quality, text rendering accuracy, and semantic alignment; details on the reward design, stochastic sampler, and training configuration are deferred to Appendix~\ref{sec:app_rl}.

The KL-divergence regularization is computed in velocity space:
\begin{equation}
D_{\text{KL}}(\pi_\theta \| \pi_{\text{ref}}) = \|\hat{v}_\theta(x_t, t) - \hat{v}_{\text{ref}}(x_t, t)\|^2.
\end{equation}

\begin{figure*}[t]
    \centering
    \includegraphics[width=0.88\linewidth]{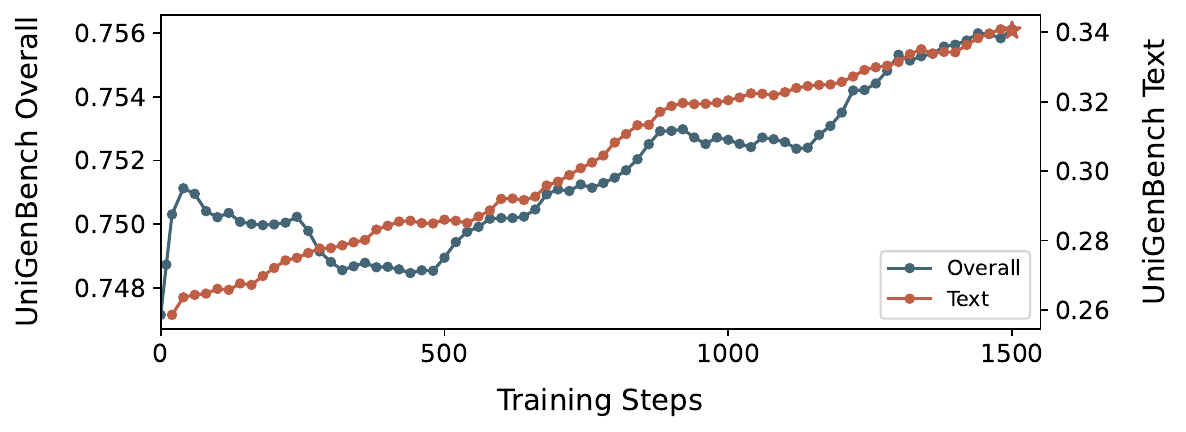} 
    \caption{UniGenBench evaluation curves during RL training over 1{,}500 steps. The left axis shows the overall score and the right axis shows the text generation sub-score. Both metrics improve steadily throughout training, with the overall score rising from $\sim$0.747 to $\sim$0.756 and the text score increasing from $\sim$0.25 to $\sim$0.34, demonstrating that RL simultaneously enhances text rendering fidelity and general generation quality.}
    \label{Fig:unigenbench_rl_eval}
\end{figure*}

While the KL penalty constrains the policy from drifting too far from the reference model, we observe that it alone is insufficient to prevent capability degradation as RL training scales beyond $\sim$1000 steps: the model exhibits a notable performance drop on tasks requiring complex instruction comprehension, such as reasoning-based generation. We attribute this to the complementary nature of the two forms of regularization: KL divergence acts as \emph{process-level guidance}, constraining the denoising trajectory to stay close to the reference policy at each step, whereas the supervised loss provides \emph{outcome-level guidance}, directly anchoring the final generation quality to the SFT distribution. Process-level constraints alone, without outcome-level anchoring, leave the model susceptible to gradual drift during prolonged training. To this end, we introduce an auxiliary supervised diffusion loss $\mathcal{L}_{\text{SFT}}$ computed on our high-quality SFT dataset, which continuously anchors the model to its supervised fine-tuning distribution. The overall training objective is:
\begin{equation}
\mathcal{L}_{\text{total}} = (1 - \lambda)\,\mathcal{L}_{\text{GRPO}} + \lambda\,\mathcal{L}_{\text{SFT}},
\end{equation}
where $\mathcal{L}_{\text{SFT}}$ is the standard flow matching loss and $\lambda$ is a small mixing coefficient. This formulation allows the model to optimize for reward signals via GRPO while retaining the generation capabilities acquired during supervised fine-tuning.

\section{Data}
The overall composition of our training data is illustrated in Fig.~\ref{Fig:data}. It combines real-world, synthetic, and carefully curated open-source datasets, covering a broad spectrum of tasks including general generation and editing, reasoning-based generation and editing, text rendering, and application-oriented scenarios.

\myparagraph{General Generation} Our pre-training corpus is sourced from several publicly available image–text pair datasets, including text-to-image-2M~\cite{text_to_image_2m_2024}, LAION-Aesthetic-6M~\cite{schuhmann2022laion5bopenlargescaledataset}, Megalith-10M~\cite{megalith10m_2024}, RedCaps-5M~\cite{desai2021redcapswebcuratedimagetextdata}, and CC-12M~\cite{changpinyo2021conceptual12mpushingwebscale}.
For high-quality instruction fine-tuning, we curate a mixture of open instruction-following datasets, including BLIP-3o (60k samples)~\cite{chen2025blip3ofamilyfullyopen}, ShareGPT-4o-Image (45k samples)~\cite{chen2025sharegpt4oimagealigningmultimodalmodels}, Echo-4o-Image (100k samples)~\cite{ye2025echo4oharnessingpowergpt4o}, and OpenGPT4o-Image (40k samples)~\cite{chen2025opengpt4oimagecomprehensivedatasetadvanced}. These are combined with 10M in-house real samples spanning both long- and short-form prompts (ratio 3:1). In addition, we synthesize approximately 50k high-clarity photorealistic images paired with fine-grained prompts using Nano Banana, further enriching detailed image generation covering both Chinese and English.

\myparagraph{General Editing} For general image editing, we collect image-instruction-image triplets from a variety of open-source datasets, including NHR-Edit~\cite{kuprashevich2025nohumansrequiredautonomoushighqualityimage} (720k samples), GPT-Image-Edit (1.5M samples)~\cite{wang2025gpt}, ShareGPT-4o-Image-Edit set (50k samples)~\cite{chen2025sharegpt4oimagealigningmultimodalmodels}, OpenGPT4o-Image-Edit set (40k samples)~\cite{chen2025opengpt4oimagecomprehensivedatasetadvanced}, Nano-banana-consist (150k samples)~\cite{nano_banana_150k}, Pico-Banana (250k samples)~\cite{qian2025picobanana400klargescaledatasettextguided}, X2I2~\cite{wu2025omnigen2explorationadvancedmultimodal}(1.6M samples) and Uniworld-Edit set~\cite{lin2025uniworldv1highresolutionsemanticencoders}(1.2M samples) together with 1.1M in-house editing samples covering both Chinese and English.

\myparagraph{Reasoning-based Generation and Editing} We utilize reasoning generation and editing datasets (150k and 100k samples, respectively) from UniReason~\cite{wang2026unireason}, covering five major knowledge domains: cultural commonsense, natural science, spatial, temporal and logical reasoning.

\begin{table*}[t]
\centering
\caption{Evaluation of reasoning-based editing involving world knowledge on the RISE~\cite{zhao2025envisioningpixelsbenchmarkingreasoninginformed} and UniREditBench~\cite{han2025unireditbench}. "*" denotes generation with textual CoT reasoning.}
\tablestyle{12pt}{1.1}
\setlength\tabcolsep{5pt}
\resizebox{1.0\textwidth}{!}{
\begin{tabular}{cccccccccc}
\toprule
\multicolumn{1}{c|}{}                                        & \multicolumn{1}{c|}{}                         & \multicolumn{5}{c|}{\bf RISE}                                                                                                                                                                   & \multicolumn{3}{c}{\bf UniREditBench}                                                         \\ \cline{3-10} 
\multicolumn{1}{c|}{\multirow{-2}{*}{\bf Model}}                 & \multicolumn{1}{c|}{\multirow{-2}{*}{\bf Params}} & \bf Temporal                    & \bf Causal                      & \bf Spatial                   & \multicolumn{1}{c|}{\bf Logical}                     & \multicolumn{1}{c|}{\bf Overall↑}                    & \bf Real World         & \multicolumn{1}{c|}{\bf Game World}         & \bf Overall↑ \\ \midrule
\rowcolor[HTML]{F6D6D3}\multicolumn{10}{c}{Closed-source Models}                                                                                                                                                                                                                                                                                                                                                              \\ \toprule
\multicolumn{1}{c|}{{\color[HTML]{1F2329} Nano Banana}}      & \multicolumn{1}{c|}{{\color[HTML]{1F2329} –}} & {\color[HTML]{1F2329} 25.9} & {\color[HTML]{1F2329} 47.8} & {\color[HTML]{1F2329} 37.0} & \multicolumn{1}{c|}{{\color[HTML]{1F2329} 18.8}} & \multicolumn{1}{c|}{{\color[HTML]{1F2329} 32.8}} & {\color[HTML]{1F2329} 75.2} & \multicolumn{1}{c|}{{\color[HTML]{1F2329} 60.4}} & 68.3     \\
\multicolumn{1}{c|}{{\color[HTML]{1F2329} GPT-Image-1}}      & \multicolumn{1}{c|}{{\color[HTML]{1F2329} –}} & {\color[HTML]{1F2329} 34.1} & {\color[HTML]{1F2329} 32.2} & {\color[HTML]{1F2329} 37.0} & \multicolumn{1}{c|}{{\color[HTML]{1F2329} 10.6}} & \multicolumn{1}{c|}{{\color[HTML]{1F2329} 28.9}} & {\color[HTML]{1F2329} 81.0}   & \multicolumn{1}{c|}{{\color[HTML]{1F2329} 62.1}} & 73.4     \\
\multicolumn{1}{c|}{{\color[HTML]{1F2329} Seedream 4.0}}     & \multicolumn{1}{c|}{{\color[HTML]{1F2329} –}} & {\color[HTML]{1F2329} 12.9} & {\color[HTML]{1F2329} 12.2} & {\color[HTML]{1F2329} 11.0} & \multicolumn{1}{c|}{{\color[HTML]{1F2329} 7.1}}  & \multicolumn{1}{c|}{{\color[HTML]{1F2329} 10.8}} & {\color[HTML]{1F2329} 66.2} & \multicolumn{1}{c|}{{\color[HTML]{1F2329} 45.4}} & 55.8     \\
\multicolumn{1}{c|}{{\color[HTML]{1F2329} FLUX-Kontext-Pro}} & \multicolumn{1}{c|}{{\color[HTML]{1F2329} –}} & {\color[HTML]{1F2329} –}    & {\color[HTML]{1F2329} –}    & {\color[HTML]{1F2329} –}  & \multicolumn{1}{c|}{{\color[HTML]{1F2329} –}}    & \multicolumn{1}{c|}{{\color[HTML]{1F2329} –}}    & {\color[HTML]{1F2329} 45.0}   & \multicolumn{1}{c|}{{\color[HTML]{1F2329} 46.5}} & 45.8     \\ \midrule
\rowcolor[HTML]{DCEBFA}\multicolumn{10}{c} {Open-source Models}                                                                                                                                                                                                                                                                                                                                         \\ \midrule
\multicolumn{1}{c|}{FLUX.1-Kontext {[}Dev{]}}                & \multicolumn{1}{c|}{12B}                      & 2.3                         & 5.5                         & 13.0                        & \multicolumn{1}{c|}{1.2}                         & \multicolumn{1}{c|}{5.8}                         & –                           & \multicolumn{1}{c|}{–}                           & –        \\
\multicolumn{1}{c|}{OmniGen2}                                & \multicolumn{1}{c|}{3B + 4B}                  & –                           & –                           & –                         & \multicolumn{1}{c|}{–}                           & \multicolumn{1}{c|}{–}                           & 53.7                        & \multicolumn{1}{c|}{33.1}                        & 43.4     \\
\multicolumn{1}{c|}{Lumina-DiMOO}                            & \multicolumn{1}{c|}{8B}                       & –                           & –                           & –                         & \multicolumn{1}{c|}{–}                           & \multicolumn{1}{c|}{–}                           & 51.4                        & \multicolumn{1}{c|}{45.6}                        & 48.5     \\
\multicolumn{1}{c|}{BAGEL\textsuperscript{*}}                                   & \multicolumn{1}{c|}{14B}                      & 5.9                         & 17.8                        & 21.0                        & \multicolumn{1}{c|}{1.2}                         & \multicolumn{1}{c|}{11.9\includegraphics[width=1em]{images/silver.png}}                        & 56.8                        & \multicolumn{1}{c|}{45.1}                        & 51.0       \\
\multicolumn{1}{c|}{Qwen-Image edit {[}2509{]}}              & \multicolumn{1}{c|}{7B + 20B}                 & 4.7                         & 10.0                          & 17.0                        & \multicolumn{1}{c|}{2.4}                         & \multicolumn{1}{c|}{8.9}                         & 71.0                          & \multicolumn{1}{c|}{41.9}                        & 56.5\includegraphics[width=1em]{images/bronze.png}     \\ \midrule
\multicolumn{1}{c|}{\bf DeepGen 1.0 (SFT)}                                 & \multicolumn{1}{c|}{\textbf{3B + 2B}}                  & 15.3                        & 18.9                        & 14.0                        & \multicolumn{1}{c|}{4.7}                         & \multicolumn{1}{c|}{13.3\includegraphics[width=1em]{images/gold.png}}                        & 74.3                        & \multicolumn{1}{c|}{80.7}                        & 77.5\includegraphics[width=1em]{images/gold.png}     \\
\multicolumn{1}{c|}{\bf DeepGen 1.0 (RL)}                              & \multicolumn{1}{c|}{\textbf{3B + 2B}}                  & 12.9                        & 14.4                        & 13.0                        & \multicolumn{1}{c|}{2.4}                         & \multicolumn{1}{c|}{10.8\includegraphics[width=1em]{images/bronze.png}}                        & 73.2                        & \multicolumn{1}{c|}{78.2}                        & 75.7\includegraphics[width=1em]{images/silver.png}     \\ \midrule
\end{tabular}}
\label{table:reason_edit}
\end{table*}

\myparagraph{Text Rendering and Application-oriented Scenarios}
To strengthen text rendering, we curate captions from document- and infographic-centric multimodal QA datasets~\cite{an2025llavaonevision15fullyopenframework}. Gemini 2.5 Pro~\cite{google_gemini25_pro_2025} is used to stochastically compose diverse rendering attributes, \textit{e.g.}, font styles, layouts, and color schemes, and combine them with an open-source prompt set tailored for text rendering from~\cite{fang2025fluxreason6mprismbenchmillionscale}. Corresponding images are synthesized using Qwen-Image, resulting in 500k text-rendering samples. We further extend the corpus to application-oriented scenarios such as Chinese poetry generation and poster design, contributing an extra 60k samples.

The detailed dataset usage in each stage is provided in Table~\ref{tab:data_details} of Appendix~\ref{pretraing_sft_details}.

\section{Experiments}

\subsection{Evaluation Setup}

\myparagraph{General Generation} We assess general text-to-image generation using GenEval~\cite{ghosh2023genevalobjectfocusedframeworkevaluating} to measure fundamental semantic alignment, and DPG-Bench~\cite{hu2024ellaequipdiffusionmodels} to assess long-prompt instruction following. In addition, we adopt UniGenBench~\cite{wang2025pref} for a comprehensive and fine-grained evaluation of general generation capability, covering ten major categories (\textit{e.g.}, attribute binding, style control, and text rendering).

\myparagraph{Reasoning Generation} We evaluate world-knowledge reasoning-based generation on WISE~\cite{niu2025wiseworldknowledgeinformedsemantic}, which contains 1,000 prompts spanning cultural knowledge, natural science, and spatial–temporal understanding. In addition, we adopt the T2I-CoREBench reasoning set~\cite{li2025easierpaintingthinkingtexttoimage}, which covers eight reasoning categories—Logical (R-LR), Behavioral (R-BR), Hypothetical (R-HR), Procedural (R-PR), Generalization (R-GR), Analogical (R-AR), Commonsense (R-CR), and Reconstructive (R-RR)—to assess reasoning generation under a structured, philosophy-inspired taxonomy.

\myparagraph{General Editing}
We evaluate general image editing on ImgEdit~\cite{ye2025imgedit} and GEdit-EN~\cite{liu2025step1x}. These benchmarks assess core editing competencies, including instruction following, editing consistency and output quality.

\myparagraph{Reasoning Editing} We evaluate world-knowledge reasoning-based image editing using UniREditBench~\cite{han2025unireditbench} with 2,700 meticulously curated samples covering both real- and game-world
scenarios, and RISE~\cite{wu2025krisbenchbenchmarkingnextlevelintelligent} with 327 samples across temporal, causal, spatial, and logical dimensions.

\myparagraph{Text Rendering}
We evaluate text rendering performance on CVTG-2K~\cite{du2025textcrafter}, which focuses on English text generation across diverse real-world scenarios, including street scenes, advertisements, and memes.

\begin{table*}[t]
\centering
\caption{Evaluation of text rendering on the CVTG-2K~\cite{du2025textcrafter}.}
\tablestyle{1pt}{1.0}
\setlength\tabcolsep{10pt}
\resizebox{0.7\textwidth}{!}{
\begin{tabular}{ccccc}
\toprule
\multicolumn{1}{c|}{\bf Model}                                  & \multicolumn{1}{c|}{\bf Params}                   & \bf Word Accuracy↑                & \bf NED↑                          & \bf CLIPScore↑                    \\ \midrule
\rowcolor[HTML]{F6D6D3}\multicolumn{5}{c}{Closed-source Models}                                                                                                                                                                    \\ \midrule
\multicolumn{1}{c|}{{\color[HTML]{1F2329} Nano Banana Pro}} & \multicolumn{1}{c|}{{\color[HTML]{1F2329} –}} & {\color[HTML]{1F2329} 0.7788} & {\color[HTML]{1F2329} 0.8754} & {\color[HTML]{1F2329} 0.7372} \\
\multicolumn{1}{c|}{{\color[HTML]{1F2329} GPT-Image-1}}     & \multicolumn{1}{c|}{{\color[HTML]{1F2329} –}} & {\color[HTML]{1F2329} 0.8569} & {\color[HTML]{1F2329} 0.9478} & {\color[HTML]{1F2329} 0.7982} \\
\multicolumn{1}{c|}{{\color[HTML]{1F2329} Seedream 4.0}}    & \multicolumn{1}{c|}{{\color[HTML]{1F2329} –}} & {\color[HTML]{1F2329} 0.8451} & {\color[HTML]{1F2329} 0.9224} & {\color[HTML]{1F2329} 0.7975} \\ \midrule
\rowcolor[HTML]{DCEBFA}\multicolumn{5}{c}{Open-source Models}                                                                                                                                             \\ \midrule
\multicolumn{1}{c|}{FLUX.1 {[}dev{]}}                       & \multicolumn{1}{c|}{12B}                      & 0.4965                        & 0.6879                        & 0.7401                        \\
\multicolumn{1}{c|}{Z-Image-Turbo}                          & \multicolumn{1}{c|}{4B + 6B}                  & 0.8585\includegraphics[width=1em]{images/bronze.png}                        & 0.9281\includegraphics[width=1em]{images/bronze.png}                        & 0.8048                        \\
\multicolumn{1}{c|}{Hunyuan-Image 3.0}                      & \multicolumn{1}{c|}{80B}                      & 0.7650                         & 0.8765                        & 0.8121\includegraphics[width=1em]{images/bronze.png}                        \\
\multicolumn{1}{c|}{Qwen-Image}                             & \multicolumn{1}{c|}{7B + 20B}                 & 0.8288                        & 0.9116                        & 0.8017                        \\
\multicolumn{1}{c|}{LongCat-Image}                          & \multicolumn{1}{c|}{7B + 6B}                  & 0.8658\includegraphics[width=1em]{images/silver.png}                        & 0.9361\includegraphics[width=1em]{images/silver.png}                         & 0.7859                        \\
\multicolumn{1}{c|}{GLM-Image}                              & \multicolumn{1}{c|}{9B + 7B}                  & 0.9116\includegraphics[width=1em]{images/gold.png}                        & 0.9557\includegraphics[width=1em]{images/gold.png}                         & 0.7877                        \\ \midrule
\multicolumn{1}{c|}{\bf DeepGen 1.0 (SFT)}                                & \multicolumn{1}{c|}{3B + 2B}                  & 0.6605                        & 0.8426                        & 0.8227\includegraphics[width=1em]{images/silver.png}                         \\
\multicolumn{1}{c|}{\bf DeepGen 1.0 (RL)}                             & \multicolumn{1}{c|}{3B + 2B}                  & 0.7533                        & 0.8936                        & 0.8278\includegraphics[width=1em]{images/gold.png}                        \\ \midrule
\end{tabular}}
\label{table:text}
\end{table*}

\subsection{Model Performance}
We compare DeepGen 1.0 against a broad set of strong baselines, covering both closed-source and open-source models. Closed-source systems include GPT-Image-1~\cite{OpenAIGPTImage1}, the Nano Banana family (i.e., Gemini-2.5-Flash-Image~\cite{google2025gemini25flashimage}), Seedream 4.0~\cite{seedream2025seedream}, and FLUX.1 Kontext [Pro]~\cite{labs2025flux}. Open-source baselines span advanced generation-only models such as FLUX.1 [Dev]~\cite{labs2025flux} and Z-Image-Turbo~\cite{cai2025z}, as well as state-of-the-art unified
multimodal models supporting both multimodal understanding and image synthesis. These include autoregressive unified models (\textit{e.g.}, Janus-Pro~\cite{chen2025janus}) and discrete diffusion-based approaches (\textit{e.g.}, Lumina-DiMOO~\cite{xin2025lumina}).

Most unified models follow the VLM–DiT paradigm, connecting VLMs with diffusion transformers via explicit connectors. Representative examples include BLIP-3o~\cite{chen2025blip3ofamilyfullyopen} and MetaQuery-XL~\cite{pan2025transfer}, which use a fixed set of learnable tokens to convey multimodal conditions to the DiT, as well as UniWorld-V1~\cite{lin2025uniworldv1highresolutionsemanticencoders}, OmniGen2~\cite{wu2025omnigen2explorationadvancedmultimodal}, the Qwen-Image series~\cite{wu2025qwenimagetechnicalreport}, and LongCat-Image~\cite{meituanlongcatteam2025longcatimagetechnicalreport}, which condition the DiT on single-layer VLM hidden states. In contrast, deep-fusion methods tightly couple VLMs and DiTs through shared attention within a unified backbone, as exemplified by Hunyuan-Image-3.0~\cite{cao2025hunyuanimage30technicalreport}, BAGEL~\cite{deng2025emergingpropertiesunifiedmultimodal}, and Show-o2~\cite{xie2025show}.

We further include models that autoregressively predicts discrete image tokens as conditions for subsequent DiT refinement, such as X-Omni~\cite{geng2025x}, GLM-Image~\cite{glm_image}, NextFlow-RL~\cite{zhang2026nextflow}, STAR~\cite{learningstar}, and Mammoth2~\cite{shen2025mammothmoda2}.
Notably, our DeepGen 1.0 remains highly lightweight, with only approximately 5B parameters, whereas most competing unified multimodal models operate at 7B parameters or more.

\subsubsection{Performance of General Generation and Editing}
As shown in Table~\ref{table:general}, DeepGen 1.0 achieves a strong performance–efficiency trade-off. With only 5B parameters (3B+2B), it consistently matches or surpasses substantially larger unified multimodal baselines across a wide range of general generation and editing benchmarks, ranking among the top three in all evaluated settings. Notably, DeepGen 1.0 unifies high-quality generation and editing within a single model, rather than relying on separate specialized models.

\myparagraph{General Generation} On GenEval~\cite{ghosh2023genevalobjectfocusedframeworkevaluating}, DeepGen 1.0 achieves 0.87, matching leading models such as Qwen-Image~\cite{wu2025qwenimagetechnicalreport} and LongCat-Image~\cite{meituanlongcatteam2025longcatimagetechnicalreport} while using significantly fewer parameters and no external LLM-based prompt rewriting. On DPGBench~\cite{hu2024ellaequipdiffusionmodels}, it scores 87.90, ranking second and demonstrating strong long-horizon instruction following ability. On the more comprehensive UniGenBench, DeepGen 1.0 achieves 75.74, again ranking second and outperforming many larger open-source baselines, including LongCat-Image~\cite{meituanlongcatteam2025longcatimagetechnicalreport}, Z-Image-Turbo~\cite{cai2025z}, and Hunyuan-Image 3.0~\cite{cao2025hunyuanimage30technicalreport}. Despite using approximately $4\times$ fewer parameters, it approaches open-source state-of-the-art performance. Overall, these results demonstrate DeepGen 1.0's robust semantic alignment, strong long-horizon instruction following for long prompts, and comprehensive fine-grained generation capabilities.

\begin{table*}[t]
\centering
\caption{Ablation study of \textbf{DeepGen 1.0} architecture.}
\small
\tablestyle{1pt}{1.0}
\setlength\tabcolsep{10pt}
\vspace{-1mm}
\resizebox{0.7\linewidth}{!}{
\begin{tabular}{c|ccccc}
\midrule
                     & GenEval& DPGBench                      & GEdit-EN & WISE & RISE \\ \midrule
DeepGen 1.0 Settings & 0.86    & \cellcolor[HTML]{FFFFFF}87.05 & 7.12  & 0.72 & 13.3 \\
w/o SCB              & 0.86    & 85.55                         & 6.75  & 0.70  & 12.6 \\
w/o Think Tokens  & 0.87    & 86.35                         & 7.02  & 0.68 & 11.7 \\
w/o Activate VLM     & 0.85    & 86.74                         & 6.93  & 0.71 & 12.9 \\ \midrule
\end{tabular}}
\vspace{-2mm}
\label{table:pretain_ab}
\end{table*}

\myparagraph{General Editing}
On ImgEdit~\cite{ye2025imgedit} and GEdit-EN~\cite{liu2025step1x}, DeepGen 1.0 remains highly competitive, ranking third under RL. It outperforms strong unified baselines such as Mammoth2, BAGEL, and OmniGen2, while approaching the performance of larger, edit-specialized models (\textit{e.g.}, Qwen-Image-Edit and LongCat-Image-Edit). Across both generation and editing, RL consistently yields further performance gains. As the RL curve on UniGenBench visualized in Fig~\ref{Fig:unigenbench_rl_eval}, RL simultaneously enhances the model’s general capabilities and text rendering performance.

\subsubsection{Performance of Reasoning-based Generation and Editing}

While maintaining strong general capabilities, DeepGen 1.0 exhibits advanced reasoning performance under a compact 5B (3B+2B) parameter budget across both reasoning-based generation and editing benchmarks. Results for world-knowledge reasoning-based generation on WISE~\cite{niu2025wiseworldknowledgeinformedsemantic}, T2I-CoREBench~\cite{li2025easierpaintingthinkingtexttoimage}, and world-knowledge-grounded editing on RISE~\cite{wu2025krisbenchbenchmarkingnextlevelintelligent} and UniREditBench~\cite{han2025unireditbench} are shown in Table~\ref{table:wise},~\ref{table:corebench}, and  ~\ref{table:reason_edit}, respectively.

\myparagraph{Reasoning-based Generation}
On WISE, DeepGen 1.0 achieves the best performance (0.73) among open-source models, outperforming strong baselines such as BAGEL~\cite{deng2025emergingpropertiesunifiedmultimodal} (relying on explicit CoT for reasoning), LongCat-Image~\cite{meituanlongcatteam2025longcatimagetechnicalreport}, and STAR~\cite{learningstar}, while further narrowing the gap to closed-source systems (\textit{e.g.}, GPT-Image-1~\cite{ OpenAIGPTImage1} and Seedream 4.0~\cite{seedream2025seedream}). Improvements are consistent across diverse knowledge domains including cultural, temporal, spatial, and natural scientific reasoning, demonstrating DeepGen 1.0's effective use of world knowledge during generation.
On T2I-CoREBench, DeepGen 1.0 attains 46.5, ranking among the top open-source models and matching or slightly surpassing substantially larger baselines such as Qwen-Image~\cite{wu2025qwenimagetechnicalreport}, Hunyuan-Image 3.0~\cite{cao2025hunyuanimage30technicalreport}, and Z-Image-Turbo~\cite{cai2025z}. This indicates broad coverage across diverse reasoning types, including logical, procedural, analogical, commonsense, and reconstructive reasoning. 

\myparagraph{Reasoning-based Editing}
DeepGen 1.0 also demonstrates strong reasoning-based editing capability. On RISE, it achieves a leading overall score 13.3 (ranked 1st) with SFT and remaining competitive under RL. On UniREditBench, it achieves 77.5 (SFT) and 75.7 (RL), significantly outperforming other open-source baselines and even exceeding the closed-source GPT-Image-1 overall. These results highlight DeepGen 1.0’s robust world-knowledge-grounded editing across both real-world and game-world scenarios~\cite{tong2025game0rl0}.

\subsubsection{Performance of Text Rendering}
As shown in Table~\ref{table:text}, DeepGen 1.0 exhibits strong text-rendering performance with only 5B parameters. RL training substantially improves Word Accuracy from 0.6605 to 0.7533, significantly enhancing character-level correctness and legibility. Meanwhile, DeepGen 1.0 preserves the highest CLIPScore (0.8278) among open-source models, indicating that improved textual fidelity does not compromise overall semantic alignment. These results validate that our RL stage effectively enhances precise text synthesis while maintaining strong instruction-level consistency.

\subsection{Ablation Study}
\subsubsection{Architecture Design}
We conduct ablation studies to quantify the contribution of key architectural components in DeepGen 1.0, by respectively implementing without applying: (1) stacked channel bridging, (2) think tokens, and (3) VLM activation.
Results across benchmarks are shown in Table~\ref{table:pretain_ab}. 

\textbf{Effect of SCB.} Removing Stacked Channel Bridging (w/o SCB) consistently degrades performance across all benchmarks: DPGBench drops from \textbf{87.05} to \textbf{85.55}, GEdit from \textbf{7.12} to \textbf{6.75}, WISE from \textbf{0.72} to \textbf{0.70}, and RISE from \textbf{13.3} to \textbf{12.6}. This verifies that SCB effectively aggregates multiple-layer VLM features and mitigates information loss compared to single-layer conditioning, thereby providing higher-quality multimodal signals to the DiT for both generation and editing. 

\begin{figure}[t]
    \centering
    \includegraphics[width=0.48\linewidth]{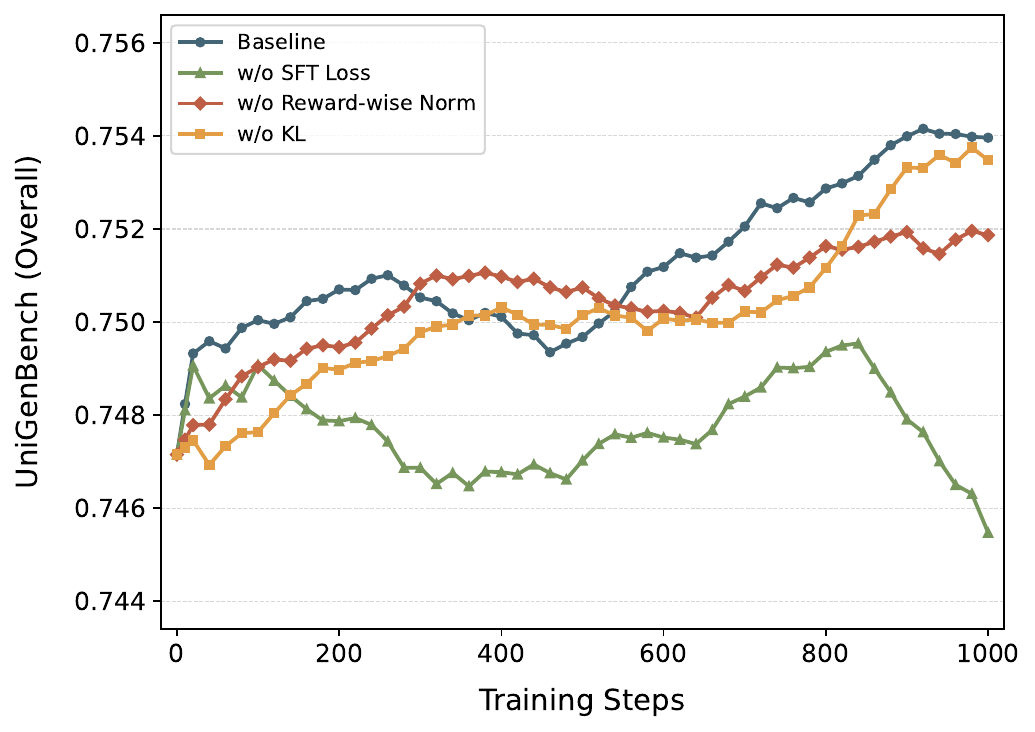}
    \hfill
    \includegraphics[width=0.48\linewidth]{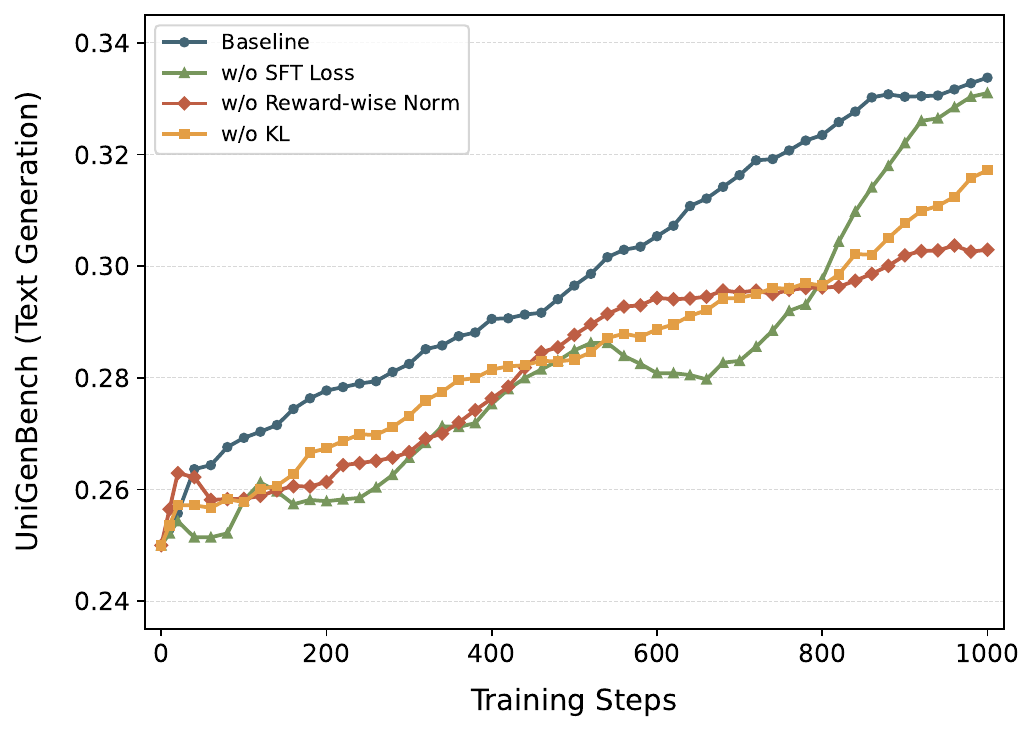}
    \caption{Evaluation curves during training for ablation variants on UniGenBench. (a) Overall score showing the importance of auxiliary SFT loss for training stability. Without it, performance degrades after $\sim$300 steps and falls well below the starting point. (b) Text generation score demonstrating that all methods improve text rendering, but removing the SFT loss results in slower and less stable progress.}
    \label{fig:ablation_curves}
\end{figure}

\textbf{Effect of Think Tokens.} Removing the learnable think tokens (w/o Think Tokens) leads to the most pronounced regression on reasoning-intensive benchmarks: WISE decreases from \textbf{0.72} to \textbf{0.68} and RISE from \textbf{13.3} to \textbf{11.7}. This suggests that think tokens serve as an implicit reasoning buffer that distills knowledge from VLM representations, strengthening world-knowledge-driven generation and editing beyond what hidden-state conditioning alone. 

\textbf{Effect of Activating the VLM.} Disabling VLM activation (w/o Activate VLM) also harms performance (\textit{e.g.}, GenEval 0.85, GEdit 6.93, WISE 0.71, RISE 12.9), indicating that modest VLM fine-tuning improves alignment with the DiT and downstream tasks, yielding more robust generation, editing, and reasoning.

\subsubsection{RL Settings}

To validate the contribution of each setting in our MR-GRPO framework, we conduct ablation studies by removing: (1) the auxiliary SFT loss, (2) the KL divergence regularization, and (3) the reward-wise advantage normalization. All variants are trained for 1{,}000 steps under identical configurations and evaluated on UniGenBench.

\myparagraph{Effect of Auxiliary SFT Loss.}
The auxiliary SFT loss is critical for maintaining generation quality during extended RL training. As shown in Figure~\ref{fig:ablation_curves}(a), removing this loss leads to performance degradation after approximately 300 steps, eventually dropping well below the initial checkpoint by the end of training. Figure~\ref{fig:ablation_curves}(b) further shows that text rendering improvement is also slower and more erratic without the SFT loss, lagging behind the baseline throughout most of training. This indicates that KL regularization alone is insufficient to anchor the model to its supervised fine-tuning distribution, and the SFT loss provides essential positive guidance that prevents capability drift and stabilizes learning across all objectives.

\myparagraph{Effect of KL Regularization.}
Removing KL regularization leads to a lower UniGenBench overall score (75.07 vs.\ 75.69) and a noticeable drop on DPGBench (87.32 vs.\ 87.75), as shown in Table~\ref{tab:ablation_rl}. Figure~\ref{fig:ablation_curves}(a) further reveals that the w/o KL variant lags behind the baseline throughout training, indicating that unconstrained policy updates can lead to forgetting of capabilities acquired during supervised fine-tuning. The combination of KL regularization and auxiliary SFT loss provides complementary constraints: KL penalizes divergence from the reference policy, while SFT loss provides positive guidance toward high-quality generation.

\myparagraph{Effect of Reward-wise Normalization.}
Normalizing advantages independently for each reward before aggregation stabilizes multi-reward optimization. As shown in Figure~\ref{fig:ablation_curves}(a), replacing reward-wise normalization with joint normalization across all rewards yields comparable performance in the early stages but leads to a growing gap after approximately 600 steps, with the final performance falling notably short of the baseline. Table~\ref{tab:ablation_rl} further shows a significant drop in text generation score (32.18 vs.\ 35.06), suggesting that high-variance rewards can dominate the policy updates and impede progress on specific objectives when normalization is not applied per reward.

\begin{table}[t]
\centering
\caption{Ablation study of RL training settings. All variants are trained for 1{,}000 steps and evaluated on generation (GenEval, DPGBench \& UniGenBench) and editing (GEdit-EN). We individually remove the auxiliary SFT loss, velocity KL regularization and reward-wise advantage normalization from the full configuration.}
\small
\tablestyle{1pt}{1.0}
\setlength\tabcolsep{10pt}
\resizebox{0.99\linewidth}{!}{%
\begin{tabular}{c|ccccc}
\toprule
                       & GenEval & DPGBench & GEdit-EN & UniGenBench (Text) & UniGenBench (Overall) \\ \midrule
DeepGen 1.0 (RL)   & \textbf{0.87}    & \textbf{87.75}    & \textbf{7.05}  & \textbf{35.06}            & \textbf{75.69}       \\
w/o Auxiliary SFT Loss & \textbf{0.87}    & 87.40 \textbf{\color{red}{(-0.35)}}     & 6.99 \textbf{\color{red}{(-0.06)}}  & 33.33 \textbf{\color{red}{(-1.73)}}             & 74.33 \textbf{\color{red}{(-1.36)}}  \\
w/o Velocity KL  & \textbf{0.87}    & 87.32 \textbf{\color{red}{(-0.43)}}    & 7.02 \textbf{\color{red}{(-0.03)}}  & 32.47 \textbf{\color{red}{(-2.59)}}             & 75.07 \textbf{\color{red}{(-0.62)}}       \\
w/o Reward-wise Norm   & 0.86 \textbf{\color{red}{(-0.01)}}    & 87.73 \textbf{\color{red}{(-0.02)}}    & 7.02 \textbf{\color{red}{(-0.03)}}  & 32.18 \textbf{\color{red}{(-2.88)}}             & 75.27 \textbf{\color{red}{(-0.42)}}       \\ \midrule
\end{tabular}
}
\label{tab:ablation_rl}
\end{table}

\section{Conclusion}

In this work, we present \modelname, a lightweight yet powerful unified multimodal model that seamlessly integrates image generation and editing within a compact 5B parameter framework. 
By synergizing a deep VLM-DiT alignment architecture with a progressive, data-centric training strategy, we demonstrate that comprehensive omni-capabilities, spanning generation, reasoning, and editing, can be achieved without relying on massive parameter scaling or excessive computational resources. 
Extensive evaluations highlight that \modelname not only outperforms existing open-source models of similar size but also rivals substantially larger systems (e.g., 80B parameters), particularly in reasoning-intensive and instruction-following tasks.

Beyond technical contributions, \modelname offers broader implications for sustainable AI. 
By decoupling high-quality generation from massive computational resources, it paves the way for accessible research on consumer-grade hardware. 
By open-sourcing \modelname, we hope it serves as a foundational step toward democratizing unified multimodal intelligence and inspiring new efficient architectures.

\bibliographystyle{unsrtnat}  
\bibliography{unireason} 

@article{deng2025emergingpropertiesunifiedmultimodal,
  title   = {Emerging Properties in Unified Multimodal Pretraining},
  author  = {Chaorui Deng and Deyao Zhu and Kunchang Li and Chenhui Gou and Feng Li and Zeyu Wang and Shu Zhong and Weihao Yu and Xiaonan Nie and Ziang Song and Guang Shi and Haoqi Fan},
  year    = {2025},
  journal = {arXiv preprint arXiv: 2505.14683}
}

@article{labs2025flux,
  title={FLUX. 1 Kontext: Flow Matching for In-Context Image Generation and Editing in Latent Space},
  author={Labs, Black Forest and Batifol, Stephen and Blattmann, Andreas and Boesel, Frederic and Consul, Saksham and Diagne, Cyril and Dockhorn, Tim and English, Jack and English, Zion and Esser, Patrick and others},
  journal={arXiv preprint arXiv:2506.15742},
  year={2025}
}

@article{niu2025wiseworldknowledgeinformedsemantic,
  title   = {WISE: A World Knowledge-Informed Semantic Evaluation for Text-to-Image Generation},
  author  = {Yuwei Niu and Munan Ning and Mengren Zheng and Weiyang Jin and Bin Lin and Peng Jin and Jiaqi Liao and Chaoran Feng and Kunpeng Ning and Bin Zhu and Li Yuan},
  year    = {2025},
  journal = {arXiv preprint arXiv: 2503.07265}
}

@article{wu2025krisbenchbenchmarkingnextlevelintelligent,
  title   = {KRIS-Bench: Benchmarking Next-Level Intelligent Image Editing Models},
  author  = {Yongliang Wu and Zonghui Li and Xinting Hu and Xinyu Ye and Xianfang Zeng and Gang Yu and Wenbo Zhu and Bernt Schiele and Ming-Hsuan Yang and Xu Yang},
  year    = {2025},
  journal = {arXiv preprint arXiv: 2505.16707}
}

@misc{google_gemini25_pro_2025,
  author       = {{Google}},
  title        = {Gemini 2.5 Pro},
  year         = {2025},
  howpublished = {\url{https://deepmind.google/models/gemini/pro/}}
}

@article{wu2025qwenimagetechnicalreport,
  title   = {Qwen-Image Technical Report},
  author  = {Chenfei Wu and Jiahao Li and Jingren Zhou and Junyang Lin and Kaiyuan Gao and Kun Yan and Sheng-ming Yin and Shuai Bai and Xiao Xu and Yilei Chen and Yuxiang Chen and Zecheng Tang and Zekai Zhang and Zhengyi Wang and An Yang and Bowen Yu and Chen Cheng and Dayiheng Liu and Deqing Li and Hang Zhang and Hao Meng and Hu Wei and Jingyuan Ni and Kai Chen and Kuan Cao and Liang Peng and Lin Qu and Minggang Wu and Peng Wang and Shuting Yu and Tingkun Wen and Wensen Feng and Xiaoxiao Xu and Yi Wang and Yichang Zhang and Yongqiang Zhu and Yujia Wu and Yuxuan Cai and Zenan Liu},
  year    = {2025},
  journal = {arXiv preprint arXiv: 2508.02324}
}

@article{chen2025sharegpt4oimagealigningmultimodalmodels,
  title   = {ShareGPT-4o-Image: Aligning Multimodal Models with GPT-4o-Level Image Generation},
  author  = {Junying Chen and Zhenyang Cai and Pengcheng Chen and Shunian Chen and Ke Ji and Xidong Wang and Yunjin Yang and Benyou Wang},
  year    = {2025},
  journal = {arXiv preprint arXiv: 2506.18095}
}

@article{zhao2025envisioningpixelsbenchmarkingreasoninginformed,
  title   = {Envisioning Beyond the Pixels: Benchmarking Reasoning-Informed Visual Editing},
  author  = {Xiangyu Zhao and Peiyuan Zhang and Kexian Tang and Xiaorong Zhu and Hao Li and Wenhao Chai and Zicheng Zhang and Renqiu Xia and Guangtao Zhai and Junchi Yan and Hua Yang and Xue Yang and Haodong Duan},
  year    = {2025},
  journal = {arXiv preprint arXiv: 2504.02826}
}

@article{li2025easierpaintingthinkingtexttoimage,
  title   = {Easier Painting Than Thinking: Can Text-to-Image Models Set the Stage, but Not Direct the Play?},
  author  = {Ouxiang Li and Yuan Wang and Xinting Hu and Huijuan Huang and Rui Chen and Jiarong Ou and Xin Tao and Pengfei Wan and Xiaojuan Qi and Fuli Feng},
  year    = {2025},
  journal = {arXiv preprint arXiv: 2509.03516}
}

@article{chen2025blip3ofamilyfullyopen,
  title   = {BLIP3-o: A Family of Fully Open Unified Multimodal Models-Architecture, Training and Dataset},
  author  = {Jiuhai Chen and Zhiyang Xu and Xichen Pan and Yushi Hu and Can Qin and Tom Goldstein and Lifu Huang and Tianyi Zhou and Saining Xie and Silvio Savarese and Le Xue and Caiming Xiong and Ran Xu},
  year    = {2025},
  journal = {arXiv preprint arXiv: 2505.09568}
}

@article{ye2025echo4oharnessingpowergpt4o,
  title={Echo-4o: Harnessing the power of gpt-4o synthetic images for improved image generation},
  author={Ye, Junyan and Jiang, Dongzhi and Wang, Zihao and Zhu, Leqi and Hu, Zhenghao and Huang, Zilong and He, Jun and Yan, Zhiyuan and Yu, Jinghua and Li, Hongsheng and others},
  journal={arXiv preprint arXiv:2508.09987},
  year={2025}
}

@article{chen2025opengpt4oimagecomprehensivedatasetadvanced,
  title={Opengpt-4o-image: A comprehensive dataset for advanced image generation and editing},
  author={Chen, Zhihong and Bai, Xuehai and Shi, Yang and Fu, Chaoyou and Zhang, Huanyu and Wang, Haotian and Sun, Xiaoyan and Zhang, Zhang and Wang, Liang and Zhang, Yuanxing and others},
  journal={arXiv preprint arXiv:2509.24900},
  year={2025}
}

@article{qian2025picobanana400klargescaledatasettextguided,
  title={Pico-banana-400k: A large-scale dataset for text-guided image editing},
  author={Qian, Yusu and Bocek-Rivele, Eli and Song, Liangchen and Tong, Jialing and Yang, Yinfei and Lu, Jiasen and Hu, Wenze and Gan, Zhe},
  journal={arXiv preprint arXiv:2510.19808},
  year={2025}
}

@misc{nano_banana_150k,
  title        = {Nano-banana-150k},
  year         = {2024},
  howpublished = {\url{https://github.com/yejy53/Nano-banana-150k}},
  note         = {GitHub repository}
}

@article{han2025unireditbench,
  title={UniREditBench: A Unified Reasoning-based Image Editing Benchmark},
  author={Han, Feng and Wang, Yibin and Li, Chenglin and Liang, Zheming and Wang, Dianyi and Jiao, Yang and Wei, Zhipeng and Gong, Chao and Jin, Cheng and Chen, Jingjing and others},
  journal={arXiv preprint arXiv:2511.01295},
  year={2025}
}

@article{ghosh2023genevalobjectfocusedframeworkevaluating,
  title={Geneval: An object-focused framework for evaluating text-to-image alignment},
  author={Ghosh, Dhruba and Hajishirzi, Hannaneh and Schmidt, Ludwig},
  journal={Advances in Neural Information Processing Systems},
  volume={36},
  pages={52132--52152},
  year={2023}
}

@article{hu2024ellaequipdiffusionmodels,
  title={Ella: Equip diffusion models with llm for enhanced semantic alignment},
  author={Hu, Xiwei and Wang, Rui and Fang, Yixiao and Fu, Bin and Cheng, Pei and Yu, Gang},
  journal={arXiv preprint arXiv:2403.05135},
  year={2024}
}

@article{wang2025pref,
  title={Pref-grpo: Pairwise preference reward-based grpo for stable text-to-image reinforcement learning},
  author={Wang, Yibin and Li, Zhimin and Zang, Yuhang and Zhou, Yujie and Bu, Jiazi and Wang, Chunyu and Lu, Qinglin and Jin, Cheng and Wang, Jiaqi},
  journal={arXiv preprint arXiv:2508.20751},
  year={2025}
}

@article{ye2025imgedit,
  title={Imgedit: A unified image editing dataset and benchmark},
  author={Ye, Yang and He, Xianyi and Li, Zongjian and Lin, Bin and Yuan, Shenghai and Yan, Zhiyuan and Hou, Bohan and Yuan, Li},
  journal={arXiv preprint arXiv:2505.20275},
  year={2025}
}

@article{liu2025step1x,
  title={Step1x-edit: A practical framework for general image editing},
  author={Liu, Shiyu and Han, Yucheng and Xing, Peng and Yin, Fukun and Wang, Rui and Cheng, Wei and Liao, Jiaqi and Wang, Yingming and Fu, Honghao and Han, Chunrui and others},
  journal={arXiv preprint arXiv:2504.17761},
  year={2025}
}

@misc{OpenAIGPTImage1,
  author = {OpenAI},
  title = {GPT-Image-1},
  year = {2025},
  url = {https://openai.com/index/introducing-4o-image-generation/},
  note = {Accessed: 2025}
}

@article{chen2025janus,
  title={Janus-pro: Unified multimodal understanding and generation with data and model scaling},
  author={Chen, Xiaokang and Wu, Zhiyu and Liu, Xingchao and Pan, Zizheng and Liu, Wen and Xie, Zhenda and Yu, Xingkai and Ruan, Chong},
  journal={arXiv preprint arXiv:2501.17811},
  year={2025}
}

@article{xin2025lumina,
  title={Lumina-dimoo: An omni diffusion large language model for multi-modal generation and understanding},
  author={Xin, Yi and Qin, Qi and Luo, Siqi and Zhu, Kaiwen and Yan, Juncheng and Tai, Yan and Lei, Jiayi and Cao, Yuewen and Wang, Keqi and Wang, Yibin and others},
  journal={arXiv preprint arXiv:2510.06308},
  year={2025}
}

@article{xie2025showosingletransformerunify,
  title={Show-o: One single transformer to unify multimodal understanding and generation},
  author={Xie, Jinheng and Mao, Weijia and Bai, Zechen and Zhang, David Junhao and Wang, Weihao and Lin, Kevin Qinghong and Gu, Yuchao and Chen, Zhijie and Yang, Zhenheng and Shou, Mike Zheng},
  journal={arXiv preprint arXiv:2408.12528},
  year={2024}
}

@article{lin2025uniworldv1highresolutionsemanticencoders,
  title={Uniworld-v1: High-resolution semantic encoders for unified visual understanding and generation. CoRR, abs/2506.03147, 2025. doi: 10. 48550},
  author={Lin, Bin and Li, Zongjian and Cheng, Xinhua and Niu, Yuwei and Ye, Yang and He, Xianyi and Yuan, Shenghai and Yu, Wangbo and Wang, Shaodong and Ge, Yunyang and others},
  journal={arXiv preprint ARXIV.2506.03147}
}

@article{wu2025omnigen2explorationadvancedmultimodal,
  title={OmniGen2: Exploration to Advanced Multimodal Generation},
  author={Wu, Chenyuan and Zheng, Pengfei and Yan, Ruiran and Xiao, Shitao and Luo, Xin and Wang, Yueze and Li, Wanli and Jiang, Xiyan and Liu, Yexin and Zhou, Junjie and others},
  journal={arXiv preprint arXiv:2506.18871},
  year={2025}
}

@article{wang2025lightfusionlightweighteddoublefusion,
  title={LightFusion: A Light-weighted, Double Fusion Framework for Unified Multimodal Understanding and Generation},
  author={Wang, Zeyu and Chen, Zilong and Gou, Chenhui and Li, Feng and Deng, Chaorui and Zhu, Deyao and Li, Kunchang and Yu, Weihao and Tu, Haoqin and Fan, Haoqi and others},
  journal={arXiv preprint arXiv:2510.22946},
  year={2025}
}

@article{cao2025hunyuanimage30technicalreport,
  title={Hunyuanimage 3.0 technical report},
  author={Cao, Siyu and Chen, Hangting and Chen, Peng and Cheng, Yiji and Cui, Yutao and Deng, Xinchi and Dong, Ying and Gong, Kipper and Gu, Tianpeng and Gu, Xiusen and others},
  journal={arXiv preprint arXiv:2509.23951},
  year={2025}
}

@article{wu2025openuni,
  title={OpenUni: A Simple Baseline for Unified Multimodal Understanding and Generation},
  author={Wu, Size and Wu, Zhonghua and Gong, Zerui and Tao, Qingyi and Jin, Sheng and Li, Qinyue and Li, Wei and Loy, Chen Change},
  journal={arXiv preprint arXiv:2505.23661},
  year={2025}
}

@article{pan2025transfer,
  title={Transfer between modalities with metaqueries},
  author={Pan, Xichen and Shukla, Satya Narayan and Singh, Aashu and Zhao, Zhuokai and Mishra, Shlok Kumar and Wang, Jialiang and Xu, Zhiyang and Chen, Jiuhai and Li, Kunpeng and Juefei-Xu, Felix and others},
  journal={arXiv preprint arXiv:2504.06256},
  year={2025}
}

@article{du2025textcrafter,
  title={Textcrafter: Accurately rendering multiple texts in complex visual scenes},
  author={Du, Nikai and Chen, Zhennan and Gao, Shan and Chen, Zhizhou and Chen, Xi and Jiang, Zhengkai and Yang, Jian and Tai, Ying},
  journal={arXiv preprint arXiv:2503.23461},
  year={2025}
}

@article{liu2025flow,
  title={Flow-grpo: Training flow matching models via online rl},
  author={Liu, Jie and Liu, Gongye and Liang, Jiajun and Li, Yangguang and Liu, Jiaheng and Wang, Xintao and Wan, Pengfei and Zhang, Di and Ouyang, Wanli},
  journal={arXiv preprint arXiv:2505.05470},
  year={2025}
}

@article{xue2025dancegrpo,
  title={DanceGRPO: Unleashing GRPO on Visual Generation},
  author={Xue, Zeyue and Wu, Jie and Gao, Yu and Kong, Fangyuan and Zhu, Lingting and Chen, Mengzhao and Liu, Zhiheng and Liu, Wei and Guo, Qiushan and Huang, Weilin and others},
  journal={arXiv preprint arXiv:2505.07818},
  year={2025}
}

@article{shao2024deepseekmath,
  title={Deepseekmath: Pushing the limits of mathematical reasoning in open language models},
  author={Shao, Zhihong and Wang, Peiyi and Zhu, Qihao and Xu, Runxin and Song, Junxiao and Bi, Xiao and Zhang, Haowei and Zhang, Mingchuan and Li, YK and Wu, Yang and others},
  journal={arXiv preprint arXiv:2402.03300},
  year={2024}
}

@article{wang2025coefficients,
  title={Coefficients-Preserving Sampling for Reinforcement Learning with Flow Matching},
  author={Wang, Feng and Yu, Zihao},
  journal={arXiv preprint arXiv:2509.05952},
  year={2025}
}

@article{liu2026gdpo,
  title={Gdpo: Group reward-decoupled normalization policy optimization for multi-reward rl optimization},
  author={Liu, Shih-Yang and Dong, Xin and Lu, Ximing and Diao, Shizhe and Belcak, Peter and Liu, Mingjie and Chen, Min-Hung and Yin, Hongxu and Wang, Yu-Chiang Frank and Cheng, Kwang-Ting and others},
  journal={arXiv preprint arXiv:2601.05242},
  year={2026}
}

@article{geng2025x,
  title={X-omni: Reinforcement learning makes discrete autoregressive image generative models great again},
  author={Geng, Zigang and Wang, Yibing and Ma, Yeyao and Li, Chen and Rao, Yongming and Gu, Shuyang and Zhong, Zhao and Lu, Qinglin and Hu, Han and Zhang, Xiaosong and others},
  journal={arXiv preprint arXiv:2507.22058},
  year={2025}
}

@article{seedream2025seedream,
  title={Seedream 4.0: Toward next-generation multimodal image generation},
  author={Seedream, Team and Chen, Yunpeng and Gao, Yu and Gong, Lixue and Guo, Meng and Guo, Qiushan and Guo, Zhiyao and Hou, Xiaoxia and Huang, Weilin and Huang, Yixuan and others},
  journal={arXiv preprint arXiv:2509.20427},
  year={2025}
}

@article{wei2026skyworkunipic20building,
  title={Skywork unipic 2.0: Building kontext model with online rl for unified multimodal model},
  author={Wei, Hongyang and Xu, Baixin and Liu, Hongbo and Wu, Size and Liu, Jie and Peng, Yi and Wang, Peiyu and Liu, Zexiang and He, Jingwen and Xietian, Yidan and others},
  journal={arXiv preprint arXiv:2509.04548},
  year={2025}
}

@article{bai2025qwen25vltechnicalreport,
  title   = {Qwen2.5-VL Technical Report},
  author  = {Shuai Bai and Keqin Chen and Xuejing Liu and Jialin Wang and Wenbin Ge and Sibo Song and Kai Dang and Peng Wang and Shijie Wang and Jun Tang and Humen Zhong and Yuanzhi Zhu and Mingkun Yang and Zhaohai Li and Jianqiang Wan and Pengfei Wang and Wei Ding and Zheren Fu and Yiheng Xu and Jiabo Ye and Xi Zhang and Tianbao Xie and Zesen Cheng and Hang Zhang and Zhibo Yang and Haiyang Xu and Junyang Lin},
  year    = {2025},
  journal = {arXiv preprint arXiv: 2502.13923}
}

@inproceedings{zhai2023sigmoidlosslanguageimage,
  title={Sigmoid loss for language image pre-training},
  author={Zhai, Xiaohua and Mustafa, Basil and Kolesnikov, Alexander and Beyer, Lucas},
  booktitle={Proceedings of the IEEE/CVF international conference on computer vision},
  pages={11975--11986},
  year={2023}
}

@inproceedings{wang2025activating,
  title={Activating Distributed Visual Region within LLMs for efficient and effective vision-language training and inference},
  author={Wang, Siyuan and Wang, Dianyi and Zhou, Chengxing and Li, Zejun and Fan, Zhihao and Huang, Xuan-Jing and Wei, Zhongyu},
  booktitle={Proceedings of the 63rd Annual Meeting of the Association for Computational Linguistics (Volume 1: Long Papers)},
  pages={30715--30727},
  year={2025}
}

@misc{text_to_image_2m_2024,
  author       = {He, Jacky and contributors},
  title        = {{text-to-image-2M}: A High-Quality, Diverse Text--Image Training Dataset},
  year         = {2024},
  howpublished = {\url{https://huggingface.co/datasets/jackyhate/text-to-image-2M}},
  note         = {Hugging Face dataset}
}

@article{schuhmann2022laion5bopenlargescaledataset,
  title={Laion-5b: An open large-scale dataset for training next generation image-text models},
  author={Schuhmann, Christoph and Beaumont, Romain and Vencu, Richard and Gordon, Cade and Wightman, Ross and Cherti, Mehdi and Coombes, Theo and Katta, Aarush and Mullis, Clayton and Wortsman, Mitchell and others},
  journal={Advances in neural information processing systems},
  volume={35},
  pages={25278--25294},
  year={2022}
}

@misc{megalith10m_2024,
  title={Team. Megalith-10M: A dataset of 10 million public-domain photographs},
  author={Matsubara, Ollin and Draw Things, AI}
}

@article{desai2021redcapswebcuratedimagetextdata,
  title={Redcaps: Web-curated image-text data created by the people, for the people},
  author={Desai, Karan and Kaul, Gaurav and Aysola, Zubin and Johnson, Justin},
  journal={arXiv preprint arXiv:2111.11431},
  year={2021}
}

@inproceedings{changpinyo2021conceptual12mpushingwebscale,
  title={Conceptual 12m: Pushing web-scale image-text pre-training to recognize long-tail visual concepts},
  author={Changpinyo, Soravit and Sharma, Piyush and Ding, Nan and Soricut, Radu},
  booktitle={Proceedings of the IEEE/CVF conference on computer vision and pattern recognition},
  pages={3558--3568},
  year={2021}
}

@article{kuprashevich2025nohumansrequiredautonomoushighqualityimage,
  title={Nohumansrequired: Autonomous high-quality image editing triplet mining},
  author={Kuprashevich, Maksim and Alekseenko, Grigorii and Tolstykh, Irina and Fedorov, Georgii and Suleimanov, Bulat and Dokholyan, Vladimir and Gordeev, Aleksandr},
  journal={arXiv preprint arXiv:2507.14119},
  year={2025}
}

@article{wang2026unireason,
  title={UniReason 1.0: A Unified Reasoning Framework for World Knowledge Aligned Image Generation and Editing},
  author={Wang, Dianyi and Ma, Chaofan and Han, Feng and Wu, Size and Song, Wei and Wang, Yibin and Zhang, Zhixiong and Wang, Tianhang and Wang, Siyuan and Wei, Zhongyu and others},
  journal={arXiv preprint arXiv:2602.02437},
  year={2026}
}

@article{fang2025fluxreason6mprismbenchmillionscale,
  title={Flux-reason-6m \& prism-bench: A million-scale text-to-image reasoning dataset and comprehensive benchmark},
  author={Fang, Rongyao and Yu, Aldrich and Duan, Chengqi and Huang, Linjiang and Bai, Shuai and Cai, Yuxuan and Wang, Kun and Liu, Si and Liu, Xihui and Li, Hongsheng},
  journal={arXiv preprint arXiv:2509.09680},
  year={2025}
}

@article{an2025llavaonevision15fullyopenframework,
  title={Llava-onevision-1.5: Fully open framework for democratized multimodal training},
  author={An, Xiang and Xie, Yin and Yang, Kaicheng and Zhang, Wenkang and Zhao, Xiuwei and Cheng, Zheng and Wang, Yirui and Xu, Songcen and Chen, Changrui and Zhu, Didi and others},
  journal={arXiv preprint arXiv:2509.23661},
  year={2025}
}

@article{meituanlongcatteam2025longcatimagetechnicalreport,
  title={Longcat-image technical report},
  author={Team, Meituan LongCat and Ma, Hanghang and Tan, Haoxian and Huang, Jiale and Wu, Junqiang and He, Jun-Yan and Gao, Lishuai and Xiao, Songlin and Wei, Xiaoming and Ma, Xiaoqi and others},
  journal={arXiv preprint arXiv:2512.07584},
  year={2025}
}

@article{li2025unifusionvisionlanguagemodelunified,
  title={UniFusion: Vision-Language Model as Unified Encoder in Image Generation},
  author={Li, Kevin and Brack, Manuel and Katakol, Sudeep and Ravi, Hareesh and Kale, Ajinkya},
  journal={arXiv preprint arXiv:2510.12789},
  year={2025}
}

@article{shi2024lmfusion,
  title={LMFusion: Adapting Pretrained Language Models for Multimodal Generation},
  author={Shi, Weijia and Han, Xiaochuang and Zhou, Chunting and Liang, Weixin and Lin, Xi Victoria and Zettlemoyer, Luke and Yu, Lili},
  journal={arXiv preprint arXiv:2412.15188},
  year={2024}
}

@article{wang2025gpt,
  title={Gpt-image-edit-1.5 m: A million-scale, gpt-generated image dataset},
  author={Wang, Yuhan and Yang, Siwei and Zhao, Bingchen and Zhang, Letian and Liu, Qing and Zhou, Yuyin and Xie, Cihang},
  journal={arXiv preprint arXiv:2507.21033},
  year={2025}
}

@article{xie2025show,
  title={Show-o2: Improved Native Unified Multimodal Models},
  author={Xie, Jinheng and Yang, Zhenheng and Shou, Mike Zheng},
  journal={arXiv preprint arXiv:2506.15564},
  year={2025}
}

@article{cai2025z,
  title={Z-image: An efficient image generation foundation model with single-stream diffusion transformer},
  author={Cai, Huanqia and Cao, Sihan and Du, Ruoyi and Gao, Peng and Hoi, Steven and Hou, Zhaohui and Huang, Shijie and Jiang, Dengyang and Jin, Xin and Li, Liangchen and others},
  journal={arXiv preprint arXiv:2511.22699},
  year={2025}
}

@article{zhang2026nextflow,
  title={NextFlow: Unified Sequential Modeling Activates Multimodal Understanding and Generation},
  author={Zhang, Huichao and Qu, Liao and Liu, Yiheng and Chen, Hang and Song, Yangyang and Dong, Yongsheng and Sun, Shikun and Li, Xian and Wang, Xu and Jiang, Yi and others},
  journal={arXiv preprint arXiv:2601.02204},
  year={2026}
}

@article{shen2025mammothmoda2,
  title={MammothModa2: A Unified AR-Diffusion Framework for Multimodal Understanding and Generation},
  author={Shen, Tao and Wan, Xin and Chen, Taicai and Zhang, Rui and Pan, Junwen and Lu, Dawei and Lei, Fanding and Lu, Zhilin and Yang, Yunfei and Cheng, Chen and others},
  journal={arXiv preprint arXiv:2511.18262},
  year={2025}
}

@article{learningstar,
  title={STAR: STacked AutoRegressive Scheme for Unified Multimodal Learning},
  author={LEARNING, UNIFIED MULTIMODAL}
}

@misc{glm_image,
  title = {GLM-Image: Auto-regressive for Dense-knowledge and High-fidelity Image Generation},
  author = {{Z.ai Team}},
  year = {2026},
  month = {jan},
  day = {14},
  url = {https://z.ai/blog/glm-image}
}

@misc{google2025gemini25flashimage,
  author = {Google},
  title = {Introducing {Gemini 2.5 Flash Image}, our state-of-the-art image model},
  howpublished = {\url{https://developers.googleblog.com/introducing-gemini-2-5-flash-image/}},
  year = {2025},
  month = {August}
}

@article{hu2022lora,
  title={Lora: Low-rank adaptation of large language models.},
  author={Hu, Edward J and Shen, Yelong and Wallis, Phillip and Allen-Zhu, Zeyuan and Li, Yuanzhi and Wang, Shean and Wang, Lu and Chen, Weizhu and others},
  journal={ICLR},
  volume={1},
  number={2},
  pages={3},
  year={2022}
}

@article{bai2025qwen30vl,
  title   = {Qwen3-VL Technical Report},
  author  = {Shuai Bai and Yuxuan Cai and Ruizhe Chen and Keqin Chen and Xionghui Chen and Zesen Cheng and Lianghao Deng and Wei Ding and Chang Gao and Chunjiang Ge and Wenbin Ge and Zhifang Guo and Qidong Huang and Jie Huang and Fei Huang and Binyuan Hui and Shutong Jiang and Zhaohai Li and Mingsheng Li and Mei Li and Kaixin Li and Zicheng Lin and Junyang Lin and Xuejing Liu and Jiawei Liu and Chenglong Liu and Yang Liu and Dayiheng Liu and Shixuan Liu and Dunjie Lu and Ruilin Luo and Chenxu Lv and Rui Men and Lingchen Meng and Xuancheng Ren and Xingzhang Ren and Sibo Song and Yuchong Sun and Jun Tang and Jianhong Tu and Jianqiang Wan and Peng Wang and Pengfei Wang and Qiuyue Wang and Yuxuan Wang and Tianbao Xie and Yiheng Xu and Haiyang Xu and Jin Xu and Zhibo Yang and Mingkun Yang and Jianxin Yang and An Yang and Bowen Yu and Fei Zhang and Hang Zhang and Xi Zhang and Bo Zheng and Humen Zhong and Jingren Zhou and Fan Zhou and Jing Zhou and Yuanzhi Zhu and Ke Zhu},
  year    = {2025},
  journal = {arXiv preprint arXiv: 2511.21631}
}

@inproceedings{ma2025janusflow,
  title={Janusflow: Harmonizing autoregression and rectified flow for unified multimodal understanding and generation},
  author={Ma, Yiyang and Liu, Xingchao and Chen, Xiaokang and Liu, Wen and Wu, Chengyue and Wu, Zhiyu and Pan, Zizheng and Xie, Zhenda and Zhang, Haowei and Yu, Xingkai and others},
  booktitle={Proceedings of the Computer Vision and Pattern Recognition Conference},
  pages={7739--7751},
  year={2025}
}

@article{cui2025paddleocr,
  title={Paddleocr 3.0 technical report},
  author={Cui, Cheng and Sun, Ting and Lin, Manhui and Gao, Tingquan and Zhang, Yubo and Liu, Jiaxuan and Wang, Xueqing and Zhang, Zelun and Zhou, Changda and Liu, Hongen and others},
  journal={arXiv preprint arXiv:2507.05595},
  year={2025}
}

@inproceedings{radford2021learning,
  title={Learning transferable visual models from natural language supervision},
  author={Radford, Alec and Kim, Jong Wook and Hallacy, Chris and Ramesh, Aditya and Goh, Gabriel and Agarwal, Sandhini and Sastry, Girish and Askell, Amanda and Mishkin, Pamela and Clark, Jack and others},
  booktitle={International conference on machine learning},
  pages={8748--8763},
  year={2021},
  organization={PmLR}
}

@article{wang2025unified,
  title={Unified multimodal chain-of-thought reward model through reinforcement fine-tuning},
  author={Wang, Yibin and Li, Zhimin and Zang, Yuhang and Wang, Chunyu and Lu, Qinglin and Jin, Cheng and Wang, Jiaqi},
  journal={arXiv preprint arXiv:2505.03318},
  year={2025}
}

@article{tong2025game0rl0,
  title   = {Game-RL: Synthesizing Multimodal Verifiable Game Data to Boost VLMs' General Reasoning},
  author  = {Jingqi Tong and Jixin Tang and Hangcheng Li and Yurong Mou and Ming Zhang and Jun Zhao and Yanbo Wen and Fan Song and Jiahao Zhan and Yuyang Lu and Chaoran Tao and Zhiyuan Guo and Jizhou Yu and Tianhao Cheng and Zhiheng Xi and Changhao Jiang and Zhangyue Yin and Yining Zheng and Weifeng Ge and Guanhua Chen and Tao Gui and Xipeng Qiu and Qi Zhang and Xuanjing Huang},
  year    = {2025},
  journal = {arXiv preprint arXiv: 2505.13886}
}

\clearpage

\beginappendix
\section{Pre-Training \& SFT Details}
\label{pretraing_sft_details}
Table~\ref{tab:data_details} and~\ref{tab:hyperparameters} provide the details of dataset usage and hyperparameter configurations at each stage, respectively.


\begin{table*}[ht]
\centering
\caption{The data details used in Pre-Training and Supervised
Fine-Tuning stages. "†" denotes covering both Chinese and English prompts.}
\label{tab:data_details}
\setlength{\tabcolsep}{8pt}            
\tablestyle{6pt}{1.2}
\begin{tabularx}{\textwidth}{c|>{\raggedright\arraybackslash}p{3.2cm}|>{\raggedright\arraybackslash}X|c}
\toprule
Stage & Task & Data source & Size \\
\midrule

Pre-Training
& \makecell[l]{General Generation}
& text-to-image-2M~\cite{text_to_image_2m_2024}, LAION-Aesthetic-6M~\cite{schuhmann2022laion5bopenlargescaledataset}, Megalith-10M~\cite{megalith10m_2024}, RedCaps-5M~\cite{desai2021redcapswebcuratedimagetextdata}, CC-12M~\cite{changpinyo2021conceptual12mpushingwebscale}
& 35M \\
\cline{2-4}
& \makecell[l]{General Editing}
& NHR-Edit~\cite{kuprashevich2025nohumansrequiredautonomoushighqualityimage}, GPT-Image-Edit~\cite{wang2025gpt}, ShareGPT-4o-Image-Edit~\cite{chen2025sharegpt4oimagealigningmultimodalmodels}, OpenGPT4o-Image-Edit~\cite{chen2025opengpt4oimagecomprehensivedatasetadvanced}, Nano-banana-consist~\cite{nano_banana_150k}, Pico-banana~\cite{qian2025picobanana400klargescaledatasettextguided}, X2I2~\cite{wu2025omnigen2explorationadvancedmultimodal}, UniWorld-Edit set~\cite{lin2025uniworldv1highresolutionsemanticencoders}, in-house editing data\textsuperscript{†}
& 6.6M \\
\midrule

\multirow{5}{*}{\makecell[c]{Supervised\\Fine-Tuning}}
& \makecell[l]{General Generation}
& BLIP-3o~\cite{chen2025blip3ofamilyfullyopen}, ShareGPT-4o-Image~\cite{chen2025sharegpt4oimagealigningmultimodalmodels}, Echo-4o-Image~\cite{ye2025echo4oharnessingpowergpt4o}, OpenGPT4o-Image~\cite{chen2025opengpt4oimagecomprehensivedatasetadvanced}, Self-Banana-50K, in-house generation data\textsuperscript{†}
& 11M \\
\cline{2-4}

& \makecell[l]{General Editing}
& NHR-Edit~\cite{kuprashevich2025nohumansrequiredautonomoushighqualityimage}, GPT-Image-Edit~\cite{wang2025gpt}, ShareGPT-4o-Image-Edit~\cite{chen2025sharegpt4oimagealigningmultimodalmodels}, OpenGPT4o-Image-Edit~\cite{chen2025opengpt4oimagecomprehensivedatasetadvanced}, Nano-banana-consist~\cite{nano_banana_150k}, Pico-banana~\cite{qian2025picobanana400klargescaledatasettextguided}, X2I2~\cite{wu2025omnigen2explorationadvancedmultimodal}, UniWorld-Edit set~\cite{lin2025uniworldv1highresolutionsemanticencoders}, in-house editing data\textsuperscript{†}
& 6.6M \\
\cline{2-4}

& \makecell[l]{Reasoning Generation}
& UniReason-T2I set~\cite{wang2026unireason}
& 150K \\
\cline{2-4}

& \makecell[l]{Reasoning Editing}
& UniReason-Edit set~\cite{wang2026unireason}
& 100K \\
\cline{2-4}

& \makecell[l]{Text Rendering}
& General text rendering, poster design\textsuperscript{†}, Chinese poem
& 560K \\
\bottomrule
\end{tabularx}
\end{table*}

\begin{table}[h]
\centering
\label{table:Hyperparameter}
\caption{Detailed Hyperparameters and Configurations of the Pre-Training and Supervised Fine-Tuning.}
\label{tab:hyperparameters}
\tablestyle{6pt}{1.2}
\begin{tabular}{l|c|c}
\toprule
\textbf{Hyperparameters} & \textbf{Stage-I (Pre-Training)} & \textbf{Stage-II (Supervised
Fine-Tuning)} \\
\midrule
Learning Rate & \(1.0 \times 10^{-4}\) & \(5.0 \times 10^{-5}\) \\

LR Scheduler & Cosine & Cosine \\

Weight Decay & 0.05 & 0.05 \\

Gradient Norm Clip & 1.0 & 1.0 \\

Optimizer & AdamW & AdamW \\

warmup ratio & 0.01 & 0.01 \\

Batch Size & 512 & 768 \\

Training GPUs & 64×H200 & 64×H200 \\
\midrule
Gen. Resolution & 512 & 512  \\

Arbitrary Resolution & x & \checkmark \\
\hline
Trainable Param & SCB connector & SCB connector, DiT, LoRA in VLM \\

LoRA Rank  & - & 64 \\
LoRA $\alpha$  & - & 128 \\
LoRA Dropout  & - & 0.05 \\
\midrule
\end{tabular}
\end{table}

\section{Reinforcement Learning Details}
\label{sec:app_rl}

\myparagraph{Noise-Preserving Stochastic Sampling.}
When sampling trajectories, the deterministic flow-matching ODE $dx_t = \hat{v}_\theta(x_t, t)\,dt$ is unsuitable for the exploration required by reinforcement learning. Prior works~\cite{liu2025flow,xue2025dancegrpo} convert it into a stochastic differential equation (SDE) to introduce randomness. However, the standard Flow-SDE formulation injects noise that exceeds the scheduler's expected noise level at each timestep, degrading sample quality and producing inaccurate reward signals. We instead adopt a noise-preserving stochastic sampling strategy~\cite{wang2025coefficients} that ensures the noise level remains consistent with the flow matching scheduler at every timestep:
\begin{equation}
x_{t-\Delta t} = \big(1 - (t\!-\!\Delta t)\big)\,\hat{x}_0 + (t\!-\!\Delta t)\cos\!\left(\tfrac{\eta\pi}{2}\right)\hat{x}_1 + (t\!-\!\Delta t)\sin\!\left(\tfrac{\eta\pi}{2}\right)\epsilon,
\end{equation}
where $\hat{x}_0 = x_t - t\,\hat{v}_\theta$ and $\hat{x}_1 = x_t + (1\!-\!t)\,\hat{v}_\theta$ are the predicted clean sample and noise respectively, $\epsilon \sim \mathcal{N}(0, I)$ is freshly sampled Gaussian noise, and $\eta \in [0, 1]$ controls the stochasticity strength. The log-probability for computing importance ratios is simplified as~\cite{wang2025coefficients}:
\begin{equation}
\log p_\theta(x_{t-\Delta t} \mid x_t) = -\|x_{t-\Delta t} - \mu_\theta(x_t, t)\|^2,
\end{equation}
where $\mu_\theta(x_t, t) = (1\!-\!(t\!-\!\Delta t))\,\hat{x}_0 + (t\!-\!\Delta t)\cos\!\left(\tfrac{\eta\pi}{2}\right)\hat{x}_1$ is the deterministic component of the sampling step. This formulation removes the variance normalization term present in the standard log-probability, avoiding numerical instability at small noise levels.

\myparagraph{Reward Functions.}
We employ three reward functions to provide complementary training signals. (1)~A VLM-based pairwise preference reward~\cite{wang2025pref} from our Unified-Reward-Think~\cite{wang2025unified} that evaluates image-text alignment and visual quality by comparing all generated images within each group and computing per-sample win rates as reward scores. (2)~An OCR reward~\cite{cui2025paddleocr} that measures text rendering accuracy by detecting rendered text in the generated image and comparing it against the target text specified in the prompt. (3)~A CLIP similarity score~\cite{radford2021learning} that captures overall semantic consistency between the generated image and the text condition. Each prompt category is assigned a different reward composition: text-rendering prompts are weighted toward the OCR reward, while general text-to-image prompts prioritize the preference reward. The detailed reward weights are provided in Table~\ref{tab:reward_weights}.

\myparagraph{Training Details.}
The RL training prompts are drawn from two categories: general text-to-image prompts and text-rendering prompts. The auxiliary SFT data is sampled from an independent curated corpus of high-quality image-text pairs covering both general generation and text rendering. Dataset details are provided below. We train with a group size of $G=8$, generating images at $512 \times 512$ resolution using 50 denoising steps. The model is optimized with a learning rate of $2 \times 10^{-6}$ for 1,500 steps. The complete set of hyperparameters is listed in Table~\ref{tab:rl_hyperparams}.

\myparagraph{Hyperparameters.}
Table~\ref{tab:rl_hyperparams} summarizes the full set of hyperparameters used for RL training.

\begin{table}[h]
\centering
\caption{Hyperparameters for reinforcement learning training.}
\label{tab:rl_hyperparams}
\tablestyle{6pt}{1.2}
\begin{tabular}{lc}
\toprule
Hyperparameter & Value \\
\midrule
Group size $G$ & 8 \\
Image resolution & $512 \times 512$ \\
Denoising steps & 50 \\
SDE stochasticity $\eta$ & 1.0 \\
Timestep fraction & 0.6 \\
Learning rate & $2 \times 10^{-6}$ \\
Total training steps & 1,500 \\
KL coefficient $\beta$ & $5 \times 10^{-7}$ \\
Clip range $\epsilon$ & $1 \times 10^{-4}$ \\
SFT auxiliary coefficient $\lambda$ & $1 \times 10^{-4}$ \\
SFT auxiliary frequency & Every step \\
Global batch size & 256 \\
DeepSpeed stage & ZeRO-2 \\
Precision & BF16 \\
\bottomrule
\end{tabular}
\end{table}

\myparagraph{Reward Weights.}
Table~\ref{tab:reward_weights} shows the per-category reward weight configuration. Text-rendering prompts are weighted toward the OCR reward to directly optimize text accuracy, while general text-to-image prompts rely primarily on the VLM-based preference reward for holistic quality assessment.

\begin{table}[h]
\centering
\caption{Reward weight configuration by prompt category.}
\label{tab:reward_weights}
\tablestyle{6pt}{1.2}
\begin{tabular}{lccc}
\toprule
Prompt Category & Preference & CLIP Sim & OCR \\
\midrule
Text rendering & 0.2 & 0.1 & 0.7 \\
General T2I & 0.7 & 0.3 & -- \\
\bottomrule
\end{tabular}
\end{table}

\myparagraph{RL Training Prompts.}
The RL training prompts consist of two categories with proportional sampling. Text-rendering prompts (sample weight 3.0$\times$) are drawn from UniGenBench text data, Qwen-Image text rendering captions, and curated text rendering prompts. General text-to-image prompts (sample weight 1.0$\times$) are sourced from UniGenBench general data, BLIP3-o captions, ShareGPT-4o image descriptions, and CoREBench prompts.

\myparagraph{Auxiliary SFT Data.}
The auxiliary supervised data for computing $\mathcal{L}_{\text{SFT}}$ is drawn from an independent corpus of high-quality image-text pairs. This corpus includes general text-to-image pairs (from BLIP3-o, ShareGPT-4o, Echo-4o, OpenGPT-4o, GenEval, and Self-Banana-50K collections) with sample weight 1.0$\times$, and text rendering pairs with sample weight 3.0$\times$ to match the emphasis on text rendering in the RL prompts.

\end{document}